\documentclass[final]{cvpr}

\usepackage{times}
\usepackage{epsfig}
\usepackage{graphicx}
\usepackage{amsmath}
\usepackage{amssymb}
\usepackage{xcolor} 
\usepackage{caption} 
\usepackage{multirow, makecell} 
\usepackage{booktabs} 
\usepackage{enumitem}
\usepackage{nopageno} 

\usepackage[pagebackref=true,breaklinks=true,colorlinks,bookmarks=false]{hyperref}

\def\cvprPaperID{6162} 
\def\confYear{CVPR 2021}

\newcommand{\model}{VITON-HD\xspace}
\newcommand{\norm}{ALIAS\xspace}

\newcommand\blfootnote[1]{
  \begingroup
  \renewcommand\thefootnote{}\footnote{#1}
  \addtocounter{footnote}{-1}
  \endgroup
}

\begin{document}

\title{VITON-HD: High-Resolution Virtual Try-On \\ via Misalignment-Aware Normalization}

\author{
Seunghwan Choi\textsuperscript{*}\quad\quad Sunghyun Park\textsuperscript{*}\quad\quad Minsoo Lee\textsuperscript{*}\quad\quad Jaegul Choo\\
KAIST, Daejeon, South Korea\\
{\tt\small \{shadow2496, psh01087, alstn2022, jchoo\}@kaist.ac.kr}
}

\twocolumn[{
\renewcommand\twocolumn[1][]{#1}
\maketitle
\begin{center}
    \centering
    \vspace{-0.9cm}
    \includegraphics[width=\linewidth]{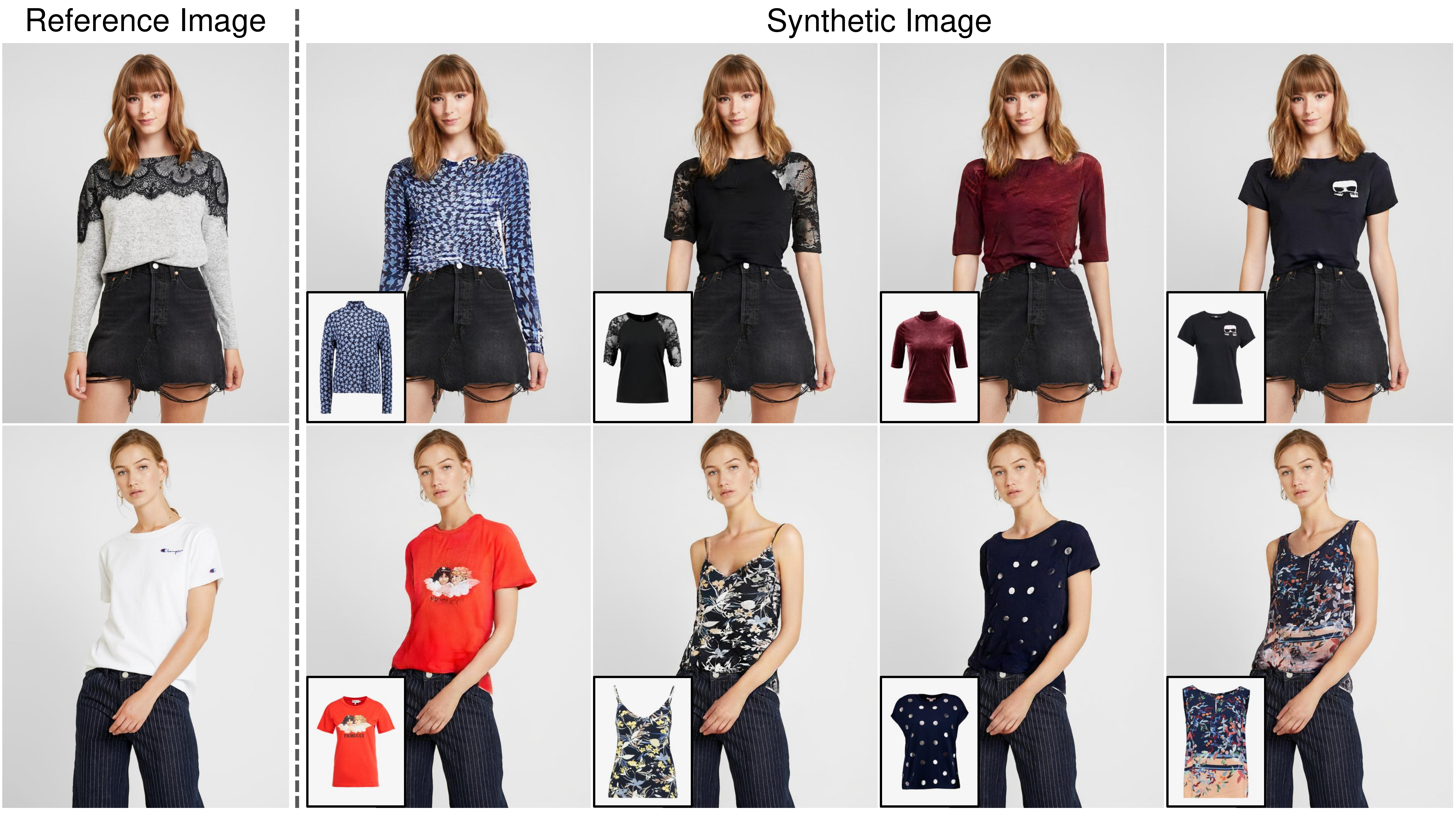}
    \vspace{-0.6cm}
    \captionof{figure}{Given a pair of a reference image (containing a person) and a target clothing image, our method successfully synthesizes 1024$\times$768 virtual try-on images.}
    \vspace{-0.3cm}
\end{center}
}]

\blfootnote{\textsuperscript{*} These authors contributed equally.}

\begin{abstract}
    The task of image-based virtual try-on aims to transfer a target clothing item onto the corresponding region of a person, which is commonly tackled by fitting the item to the desired body part and fusing the warped item with the person.
    While an increasing number of studies have been conducted, the resolution of synthesized images is still limited to low (\eg, 256$\times$192), which acts as the critical limitation against satisfying online consumers.
    We argue that the limitation stems from several challenges:
    as the resolution increases, the artifacts in the misaligned areas between the warped clothes and the desired clothing regions become noticeable in the final results;
    the architectures used in existing methods have low performance in generating high-quality body parts and maintaining the texture sharpness of the clothes.
    To address the challenges, we propose a novel virtual try-on method called \model that successfully synthesizes 1024$\times$768 virtual try-on images.
    Specifically, we first prepare the segmentation map to guide our virtual try-on synthesis, and then roughly fit the target clothing item to a given person's body.
    Next, we propose ALIgnment-Aware Segment (\norm) normalization and \norm generator to handle the misaligned areas and preserve the details of 1024$\times$768 inputs.
    Through rigorous comparison with existing methods, we demonstrate that \model highly surpasses the baselines in terms of synthesized image quality both qualitatively and quantitatively.
    Code is available at \url{https://github.com/shadow2496/VITON-HD}.
\end{abstract}

\vspace{-0.2cm}
\section{Introduction}

Image-based virtual try-on refers to the image generation task of changing the clothing item on a person into a different item, given in a separate product image.
With a growing trend toward online shopping, virtually wearing the clothes can enrich a customer's experience, as it gives an idea about how these items would look on them.

Virtual try-on is similar to image synthesis, but it has unique and challenging aspects.
Given images of a person and a clothing product, the synthetic image should meet the following criteria:
(1) The person's pose, body shape, and identity should be preserved.
(2) The clothing product should be naturally deformed to the desired clothing region of the given person, by reflecting his/her pose and body shape.
(3) Details of the clothing product should be kept intact.
(4) The body parts initially occluded by the person's clothes in the original image should be properly rendered.
Since the given clothing image is not initially fitted to the person image, fulfilling these requirements is challenging, which leaves the development of virtual try-on still far behind the expectations of online consumers.
In particular, the resolution of virtual try-on images is low compared to the one of normal pictures on online shopping websites.

After Han \etal~\cite{han2018viton} proposed VITON, various image-based virtual try-on methods have been proposed~\cite{wang2018toward, yu2019vtnfp, yang2020towards, dong2019fw}.
These methods follow two processes in common:
(1) warping the clothing image initially to fit the human body;
(2) fusing the warped clothing image and the image of the person that includes pixel-level refinement.
Also, several recent methods~\cite{han2019clothflow, yu2019vtnfp, yang2020towards} add a module that generates segmentation maps and determine the person's layout from the final image in advance.

\begin{figure}[t!]
    \centering
    \includegraphics[width=1.0\linewidth]{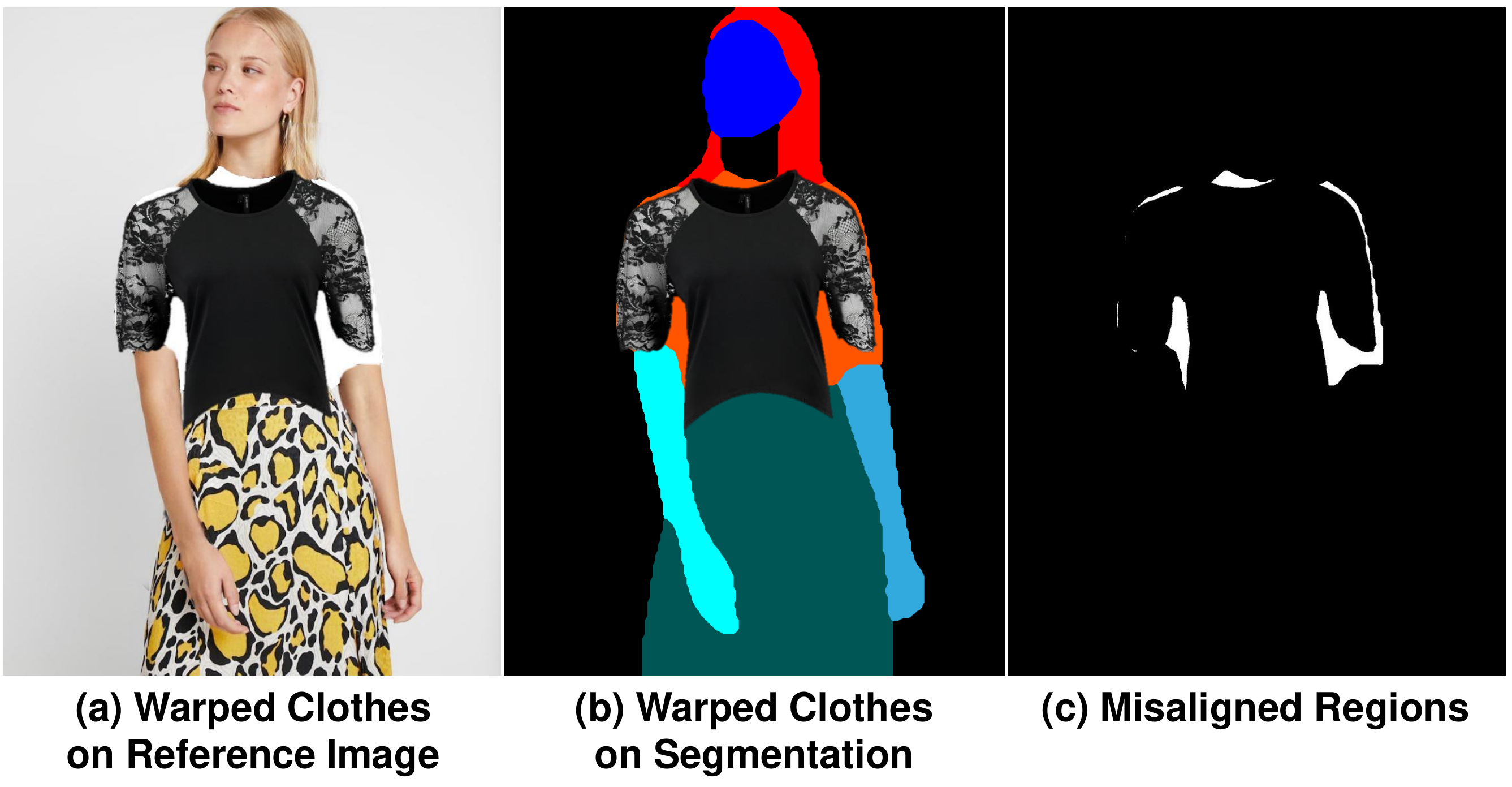}
    \vspace{-0.7cm}
    \caption{An example of misaligned regions.}
    \label{fig:misaligned regions}
    \vspace{-0.5cm}
\end{figure}

However, the resolution of the synthetic images from the previous methods is low (\eg, 256$\times$192) due to the following reasons.
First, the misalignment between the warped clothes and a person's body results in the artifacts in the misaligned regions, which become noticeable as the image size increases.
It is difficult to warp clothing images to fit the body perfectly, so the misalignment occurs as shown in Fig.~\ref{fig:misaligned regions}.
Most of previous approaches utilize the thin-plate spline (TPS) transformation to deform clothing images.
To accurately deform clothes, ClothFlow~\cite{han2019clothflow} predicts the optical flow maps of the clothes and the desired clothing regions.
However, the optical flow maps does not remove the misalignment completely on account of the regularization.
In addition, the process requires more computational costs than other methods due to the need of predicting the movement of clothes at a pixel level.
(The detailed analysis of ClothFlow is included in the supplementary.)
Second, a simple U-Net architecture~\cite{ronneberger2015u} used in existing approaches is insufficient in synthesizing initially occluded body parts in final high-resolution (\eg, 1024$\times$768) images.
As noted in Wang \etal~\cite{wang2018high}, applying a simple U-Net-based architecture to generate high-resolution images leads to unstable training as well as unsatisfactory quality of generated images.
Also, refining the images once at the pixel level is insufficient in preserving the details of high-resolution clothing images.

To address the above-mentioned challenges, we propose a novel high-resolution virtual try-on method, called \model.
In particular, we introduce a new clothing-agnostic person representation that leverages the pose information and the segmentation map so that the clothing information is eliminated thoroughly.
Afterwards, we feed the segmentation map and the clothing item deformed to fit the given human body to the model.
Using the additional information, our novel ALIgnment-Aware Segment (\norm) normalization removes information irrelevant to the clothing texture in the misaligned regions and propagates the semantic information throughout the network.
The normalization separately standardizes the activations corresponding to the misaligned regions and the other regions, and modulates the standardized activations using the segmentation map.
Our \norm generator employing \norm normalization synthesizes the person image wearing the target product while filling the misaligned regions with the clothing texture and preserving the details of the clothing item through the multi-scale refinement at a feature level.
To validate the performance of our framework, we collected a 1024$\times$768 dataset that consists of pairs of a person and a clothing item for our research purpose.
Our experiments demonstrate that \model significantly outperforms the existing methods in generating 1024$\times$768 images, both quantitatively and qualitatively.
We also confirm the superior capability of our novel \norm normalization module in dealing with the misaligned regions.

We summarize our contributions as follows:
\begin{itemize}[noitemsep,topsep=0pt]
    \item We propose a novel image-based virtual try-on approach called \model, which is, to the best of our knowledge, the first model to successfully synthesize 1024$\times$768 images.
    \item We introduce a clothing-agnostic person representation that allows our model to remove the dependency on the clothing item originally worn by the person.
    \item To address the misalignment between the warped clothes and the desired clothing regions, we propose \norm normalization and \norm generator, which is effective in maintaining the details of clothes.
    \item We demonstrate the superior performance of our method through experiments with baselines on the newly collected dataset.
\end{itemize}

\section{Related Work}

\textbf{Conditional Image Synthesis.}
Conditional generative adversarial networks (cGANs) utilize additional information, such as class labels~\cite{odena2017conditional, brock2018large}, text~\cite{reed2016generative, xu2018attngan}, and attributes~\cite{shen2017learning}, to steer the image generation process.
Since the emergence of pix2pix~\cite{isola2017image}, numerous cGANs conditioned on input images have been proposed to generate high-resolution images in a stable manner~\cite{wang2018high, anokhin2020high, park2020swapping}.
However, these methods tend to generate blurry images when handling a large spatial deformation between the input image and the target image.
In this paper, we propose a method that can address the spatial deformation of input images and properly generate 1024$\times$768 images.

\textbf{Normalization Layers.}
Normalization layers~\cite{ioffe2015batch, ulyanov2016instance} have been widely applied in modern deep neural networks.
Normalization layers, whose affine parameters are estimated with external data, are called conditional normalization layers.
Conditional batch normalization~\cite{de2017modulating} and adaptive instance normalization~\cite{huang2017arbitrary} are such conditional normalization techniques and have been used in style transfer tasks.
SPADE~\cite{park2019semantic} and SEAN~\cite{zhu2020sean} utilize segmentation maps to apply spatially varying affine transformations.
Using the misalignment mask as external data, our proposed normalization layer computes the means and the variances of the misaligned area and the other area within an instance separately.
After standardization, we modulate standardized activation maps with affine parameters inferred from human-parsing maps to preserve semantic information.

\textbf{Virtual Try-On Approaches.}
There are two main categories for virtual try-on approaches: 3D model-based approaches~\cite{guan2012drape, sekine2014virtual, pons2017clothcap, patel2020tailornet} and 2D image-based approaches~\cite{han2018viton, wang2018toward, han2019clothflow, yu2019vtnfp, yang2020towards, dong2019towards}.
3D model-based approaches can accurately simulate the clothes but are not widely applicable due to their dependency on 3D measurement data.

2D image-based approaches do not rely on any 3D information, thus being computationally efficient and appropriate for practical use.
Jetchev and Bergmann~\cite{jetchev2017conditional} proposed CAGAN, which first introduced the task of swapping fashion articles on human images.
VITON~\cite{han2018viton} addressed the same problem by proposing a coarse-to-fine synthesis framework that involves TPS transformation of clothes.
Most existing virtual try-on methods tackle different aspects of VITON to synthesize perceptually convincing photo-realistic images.
CP-VTON~\cite{wang2018toward} adopted a geometric matching module to learn the parameters of TPS transformation, which improves the accuracy of deformation.
VTNFP~\cite{yu2019vtnfp} and ACGPN~\cite{yang2020towards} predicted the human-parsing maps of a person wearing the target clothes in advance to guide the try-on image synthesis.
Even though the image quality at high resolution is an essential factor in evaluating the practicality of the generated images, none of the methods listed above could generate such photo-realistic images at high resolution.

\begin{figure*}[t!]
    \centering
    \includegraphics[width=1.0\linewidth]{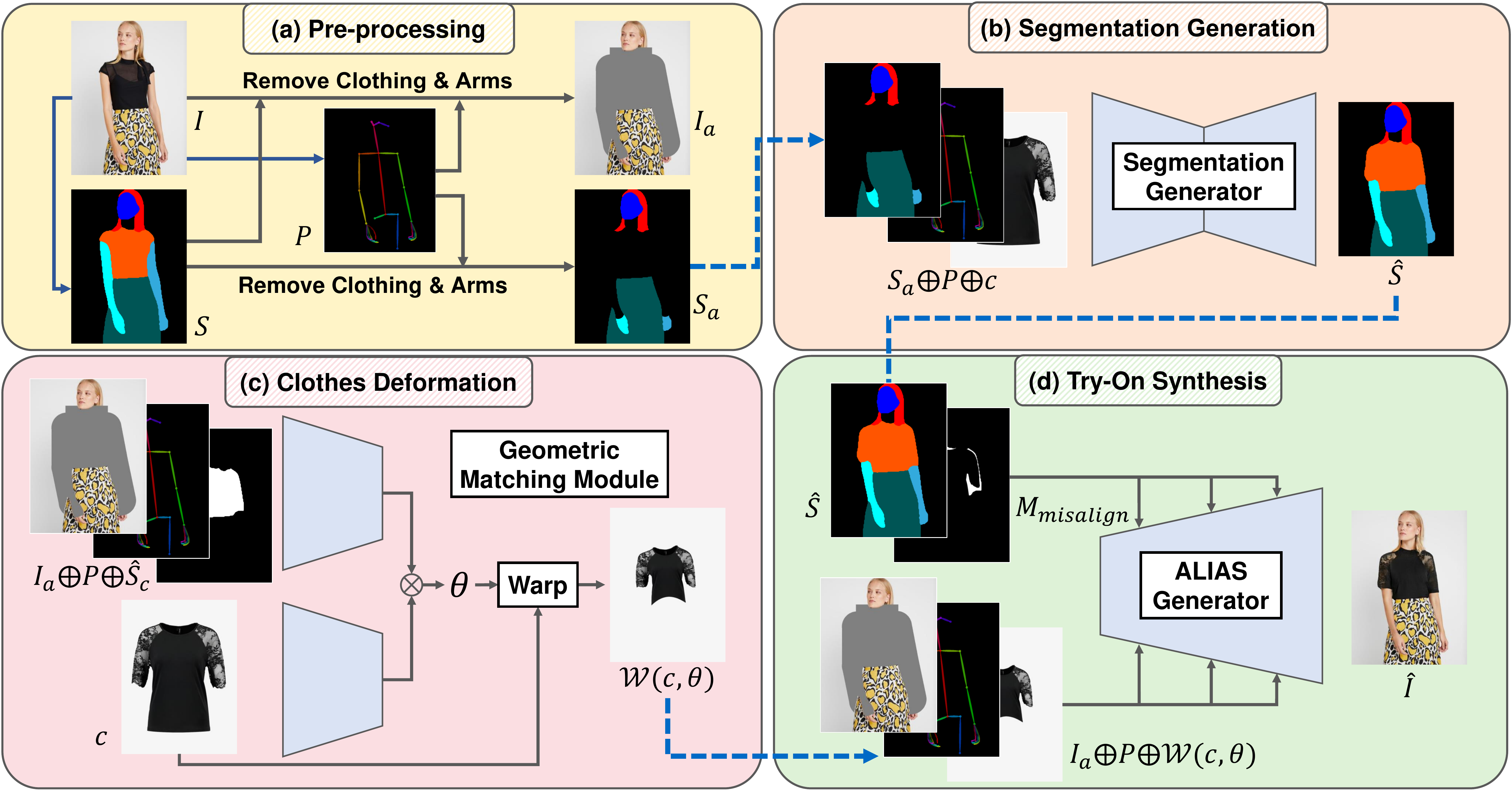}
    \vspace{-0.5cm}
    \caption{Overview of a \model. (a) First, given a reference image $I$ containing a target person, we predict the segmentation map $S$ and the pose map $P$, and utilize them to pre-process $I$ and $S$ as a clothing-agnostic person image $I_a$ and segmentation $S_a$. (b) Segmentation generator produces the synthetic segmentation $\hat{S}$ from $(S_a, P, c)$. (c) Geometric matching module deforms the clothing image $c$ according to the predicted clothing segmentation $\hat{S_c}$ extracted from $\hat{S}$. (d) Finally, \norm generator synthesizes the final output image $\hat{I}$ based on the outputs from the previous stages via our \norm normalization.}
    \vspace{-0.4cm}
    \label{fig:overview}
\end{figure*}

\section{Proposed Method}
\textbf{Model Overview.}
As described in Fig.~\ref{fig:overview}, given a reference image $I \in \mathbb{R}^{3 \times H \times W}$ of a person and a clothing image $c \in \mathbb{R}^{3 \times H \times W}$ ($H$ and $W$ denote the image height and width, respectively), the goal of \model is to generate a synthetic image $\hat{I} \in \mathbb{R}^{3 \times H \times W}$ of the same person wearing the target clothes $c$, where the pose and body shape of $I$ and the details of $c$ are preserved.
While training the model with $(I, c, \hat{I})$ triplets is straightforward, construction of such dataset is costly.
Instead, we use $(I, c, I)$ where the person in the reference image $I$ is already wearing $c$.

Since directly training on $(I, c, I)$ can harm the model's generalization ability at test time, we first compose a clothing-agnostic person representation that leaves out the clothing information in $I$ and use it as an input.
Our new clothing-agnostic person representation uses both the pose map and the segmentation map of the person to eliminate the clothing information in $I$ (Section~\ref{sec:person representation}).
The model generates the segmentation map from the clothing-agnostic person representation to help the generation of $\hat{I}$ (Section~\ref{sec:segmentation}).
We then deform $c$ to roughly align it to the human body (Section~\ref{sec:deformation}).
Lastly, we propose the ALIgnment-Aware Segment (ALIAS) normalization that removes the misleading information in the misaligned area after deforming $c$.
\norm generator fills the misaligned area with the clothing texture and maintains the clothing details (Section~\ref{sec:try-on synthesis}).

\subsection{Clothing-Agnostic Person Representation}\label{sec:person representation}
To train the model with pairs of $c$ and $I$ already wearing $c$, a person representation without the clothing information in $I$ has been utilized in the virtual try-on task.
Such representations have to satisfy the following conditions:
(1) the original clothing item to be replaced should be deleted;
(2) sufficient information to predict the pose and the body shape of the person should be maintained;
(3) the regions to be preserved (\eg, face and hands) should be kept to maintain the person's identity.

\textbf{Problems of Existing Person Representations.}
In order to maintain the person's shape, several approaches~\cite{han2018viton, wang2018toward, yu2019vtnfp} provide a coarse body shape mask as a cue to synthesize the image, but fail to reproduce the body parts elaborately (\eg, hands).
To tackle this issue, ACGPN~\cite{yang2020towards} employs the detailed body shape mask as the input, and the neural network attempts to discard the clothing information to be replaced.
However, since the body shape mask includes the shape of the clothing item, neither the coarse body shape mask nor the neural network could perfectly eliminate the clothing information.
As a result, the original clothing item that is not completely removed causes problems in the test phase.

\textbf{Clothing-Agnostic Person Representation.}
We propose a clothing-agnostic image $I_a$ and a clothing-agnostic segmentation map $S_a$ as inputs of each stage, which truly eliminate the shape of clothing item and preserve the body parts that need to be reproduced.
We first predict the segmentation map $S \in \mathbb{L}^{H \times W}$ and the pose map $P \in \mathbb{R}^{3 \times H \times W}$ of the image $I$ by utilizing the pre-trained networks~\cite{gong2018instance, cao2019openpose}, where $\mathbb{L}$ is a set of integers indicating the semantic labels.
The segmentation map $S$ is used to remove the clothing region to be replaced and preserve the rest of the image.
The pose map $P$ is utilized to remove the arms, but not the hands, as they are difficult to reproduce.
Based on $S$ and $P$, we generate the clothing-agnostic image $I_a$ and the clothing-agnostic segmentation map $S_a$, which allow the model to remove the original clothing information thoroughly, and preserve the rest of the image.
In addition, unlike other previous work, which adopts the pose heatmap with each channel corresponded to one keypoint, we concatenate $I_a$ or $S_a$ to the RGB pose map $P$ representing a skeletal structure that improves generation quality.

\begin{figure*}[t!]
    \centering
    \includegraphics[width=1.0\linewidth]{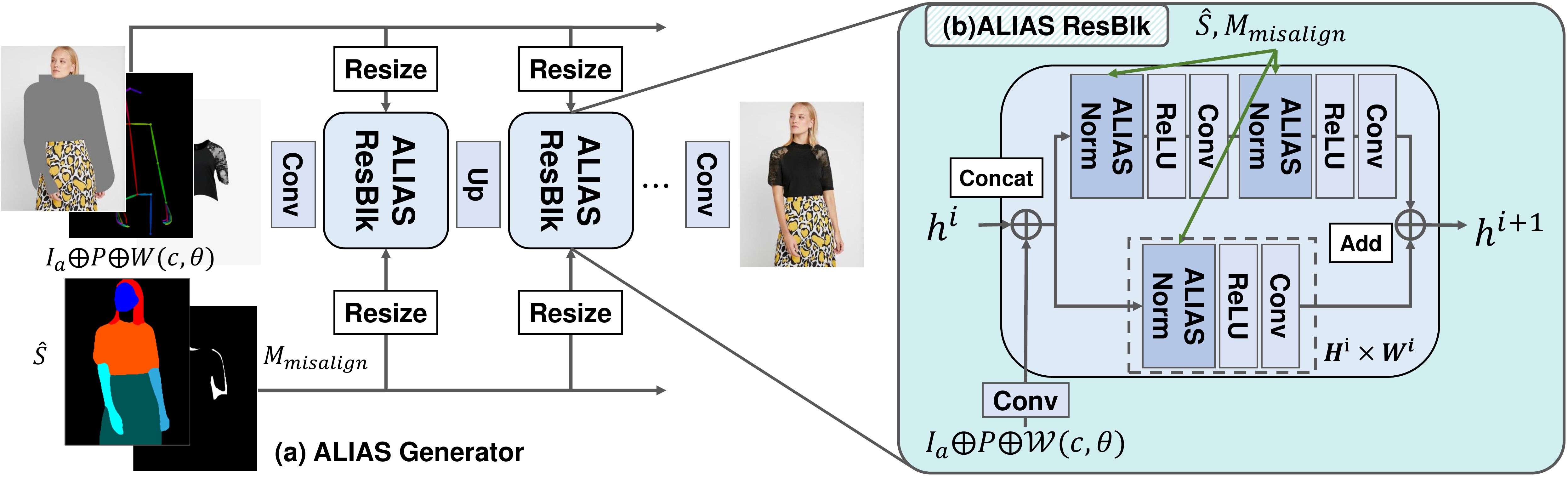}
    \vspace{-0.5cm}
    \caption{\norm generator.
    (a) The \norm generator is composed of a series of \norm residual blocks, along with upsampling layers.
    The input $(I_a, P, \mathcal{W}(c, \theta))$ is resized and injected into each layer of the generator.
    (b) A detailed view of a \norm residual block.
    Resized $(I_a, P, \mathcal{W}(c,\theta))$ is concatenated to $h^{i}$ after passing through a convolution layer.
    Each \norm normalization layer leverages resized $\hat{S}$ and $M_{misalign}$ to normalize the activation.
    }
    \vspace{-0.4cm}
    \label{fig:try-on synthesis}
\end{figure*}

\subsection{Segmentation Generation}\label{sec:segmentation}
Given the clothing-agnostic person representation $(S_a, P)$, and the target clothing item $c$, the segmentation generator $G_S$ predicts the segmentation map $\hat{S} \in \mathbb{L}^{H \times W}$ of the person in the reference image wearing $c$.
We train $G_S$ to learn the mapping between $S$ and $(S_a, P, c)$, in which the original clothing item information is completely removed.
As the architecture of $G_S$, we adopt U-Net~\cite{ronneberger2015u}, and the total loss $\mathcal{L}_S$ of the segmentation generator are written as 
\begin{equation}
    \mathcal{L}_{S} = \mathcal{L}_{cGAN} + \lambda_{CE} \mathcal{L}_{CE},
    \label{eq:segloss}
\end{equation}
where $\mathcal{L}_{CE}$ and $\mathcal{L}_{cGAN}$ denote the pixel-wise cross-entropy loss and conditional adversarial loss between $\hat{S}$ and $S$, respectively. $\lambda_{CE}$ is the hyperparameter corresponding to the relative importance between two losses.

\subsection{Clothing Image Deformation}\label{sec:deformation}
In this stage, we deform the target clothing item $c$ to align it with $\hat{S}_c$, which is the clothing area of $\hat{S}$.
We employ the geometric matching module proposed in CP-VTON~\cite{wang2018toward} with the clothing-agnostic person representation $(I_a, P)$ and $\hat{S}_c$ as inputs.
A correlation matrix between the features extracted from $(I_a, P)$ and $c$ is first calculated .
With the correlation matrix as an input, the regression network predicts the TPS transformation parameters $\theta \in \mathbb{R}^{2 \times 5 \times 5}$, and then $c$ is warped by $\theta$.
In the training phase, the model takes $S_c$ extracted from $S$ instead of $\hat{S}_c$.
The module is trained with the L1 loss between the warped clothes and the clothes $I_c$ that is extracted from $I$.
In addition, the second-order difference constraint~\cite{yang2020towards} is adopted to reduce obvious distortions in the warped clothing images from deformation.
The overall objective function to warp the clothes to fit the human body is written as
\begin{equation}
    \mathcal{L}_{warp} = ||I_{c} - \mathcal{W}(c, \theta)||_{1,1}
    + \lambda_{const} \mathcal{L}_{const},
\end{equation}
where $\mathcal{W}$ is the function that deforms $c$ using $\theta$, $\mathcal{L}_{const}$ is a second-order difference constraint, and $\lambda_{const}$ is the hyperparameter for $\mathcal{L}_{const}$.

\subsection{Try-On Synthesis via \norm Normalization}\label{sec:try-on synthesis}
We aim to generate the final synthetic image $\hat{I}$ based on the outputs from the previous stages.
Overall, we fuse the clothing-agnostic person representation $(I_a, P)$ and the warped clothing image $\mathcal{W}(c, \theta)$, guided by $\hat{S}$.
$(I_a, P, \mathcal{W}(c, \theta))$ is injected into each layer of the generator.
For $\hat{S}$, we propose a new conditional normalization method called the ALIgnment-Aware Segment (\norm) normalization.
\norm normalization enables the preservation of semantic information, and the removal of misleading information from the misaligned regions by leveraging $\hat{S}$ and the mask of these regions.

\begin{figure}[t!]
    \centering
    \includegraphics[width=1.0\linewidth]{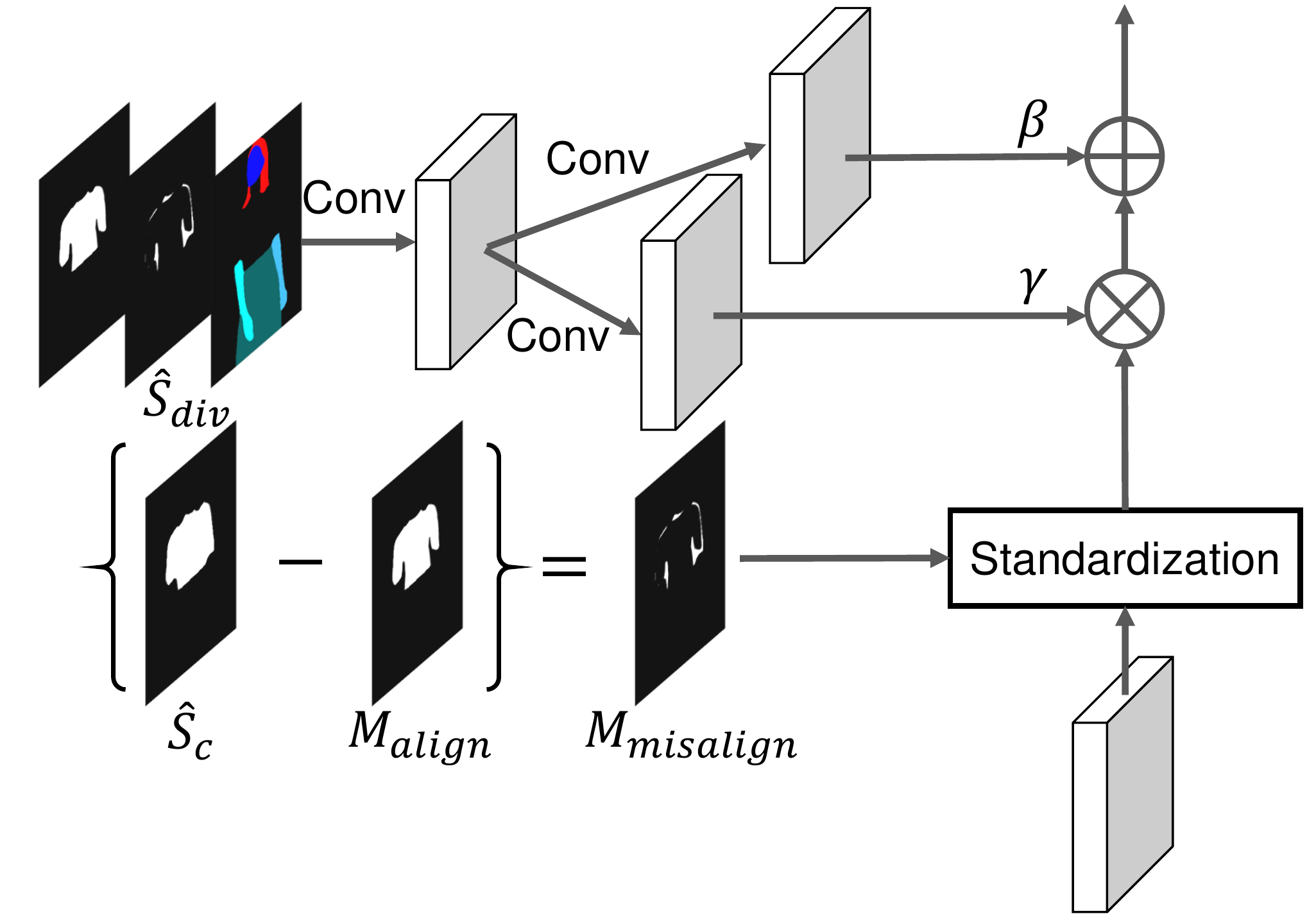}
    \vspace{-0.5cm}
    \caption{\norm normalization.
    First, the activation is separately standardized according to the regions divided by $M_{misalign}$, which can be obtained from the difference between $\hat{S}_c$ and $M_{align}$.
    Next, $\hat{S}_{div}$ is convolved to create the modulation parameters $\gamma$ and $\beta$, and then the standardized activation is modulated with the parameters $\gamma$ and $\beta$.}
    \vspace{-0.4cm}
    \label{fig:alias norm}
\end{figure}

\begin{table*}[t!]
\centering
\small
    \begin{tabular}{@{}p{0.10\textwidth} p{0.04\textwidth}<{\centering} p{0.04\textwidth}<{\centering} p{0.04\textwidth}<{\centering} | p{0.04\textwidth}<{\centering} p{0.04\textwidth}<{\centering} p{0.04\textwidth}<{\centering} | p{0.04\textwidth}<{\centering} p{0.04\textwidth}<{\centering} p{0.04\textwidth}<{\centering}@{}}
    \toprule
    & \multicolumn{3}{c}{256 $\times$ 192} & \multicolumn{3}{c}{512 $\times$ 384} & \multicolumn{3}{c}{1024 $\times$ 768} \\
    & SSIM$_{\uparrow}$ & LPIPS$_{\downarrow}$ & FID$_{\downarrow}$ & SSIM$_{\uparrow}$ & LPIPS$_{\downarrow}$ & FID$_{\downarrow}$ & SSIM$_{\uparrow}$ & LPIPS$_{\downarrow}$ & FID$_{\downarrow}$ \\
    \midrule
    CP-VTON & 0.739 & 0.159 & 56.23 & 0.791 & 0.141 & 31.96 & 0.786 & 0.158 & 43.28 \\
    ACGPN & 0.842 & 0.064 & \textbf{26.45} & 0.863 & 0.067 & 15.22 & 0.856 & 0.102 & 43.39 \\
    \midrule
    VITON-HD*  & - & - & - & - & - & - & 0.893 & 0.054 & 12.47 \\
    VITON-HD & \textbf{0.844} & \textbf{0.062} & 27.83 & \textbf{0.870} & \textbf{0.052} & \textbf{14.05} & \textbf{0.895} & \textbf{0.053} & \textbf{11.74} \\
    \bottomrule
    \end{tabular}
    \vspace{-0.2cm}
    \caption{Quantitative comparison with baselines across different resolutions. 
    VITON-HD* is a \model variant where the standardization in \norm normalization is replaced by channel-wise standardization as in the original instance normalization.
    For the SSIM, higher is better. For the LPIPS and the FID, lower is better.}
    \label{Table:resolution comparison}
    \vspace{-0.4cm}
\end{table*}

\textbf{Alignment-Aware Segment Normalization.}
Let us denote $h^i \in \mathbb{R}^{N \times C^i \times H^i \times W^i}$ as the activation of the $i$-th layer of a network for a batch of $N$ samples, where $H^i$, $W^i$, and $C^i$ indicate the height, width, and the number of channels of $h^i$, respectively. 
\norm normalization has two inputs:
(1) the synthetic segmentation map $\hat{S}$;
(2) the misalignment binary mask $M_{misalign} \in \mathbb{L}^{H \times W}$, which excludes the warped mask of the target clothing image $\mathcal{W}(M_c, \theta)$ from $\hat{S}_c$ ($M_c$ denotes the target clothing mask), \ie,
\begin{equation}
    M_{align} = \hat{S}_c \cap \mathcal{W}(M_c, \theta)
    \label{eq:alignment mask}
\end{equation}
\begin{equation}
    M_{misalign} = \hat{S}_c - M_{align}.
    \label{eq:misalignment mask}
\end{equation}

Fig.~\ref{fig:alias norm} illustrates the workflow of the \norm normalization.
We first obtain $M_{align}$ and $M_{misalign}$ from Eq.~\eqref{eq:alignment mask} and Eq.~\eqref{eq:misalignment mask}.
We define the modified version of $\hat{S}$ as $\hat{S}_{div}$, where $\hat{S}_c$ in $\hat{S}$ separates into $M_{align}$ and $M_{misalign}$.
\norm normalization standardizes the regions of $M_{misalign}$ and the other regions in $h^i$ separately, and then modulates the standardized activation using affine transformation parameters inferred from $\hat{S}_{div}$.
The activation value at site ($n \in N, k \in C^i, y \in H^i, x \in W^i$) is calculated by
\begin{equation}
    \gamma^i_{k,y,x}(\hat{S}_{div})
    {{h^i_{n,k,y,x} - \mu^{i,m}_{n,k}} \over \sigma^{i,m}_{n,k}}
    + \beta^i_{k,y,x}(\hat{S}_{div}), 
\end{equation}
where $h^i_{n,k,y,x}$ is the activation at the site before normalization and $\gamma^i_{k,y,x}$ and $\beta^i_{k,y,x}$ are the functions that convert $\hat{S}_{div}$ to modulation parameters of the normalization layer.
$\mu^{i,m}_{n,k}$ and $\sigma^{i,m}_{n,k}$ are the mean and standard deviation of the activation in sample $n$ and channel $k$.
$\mu^{i,m}_{n,k}$ and $\sigma^{i,m}_{n,k}$ are calculated by
\begin{equation}
    \mu^{i,m}_{n,k} = {1 \over |\Omega^{i,m}_n|}
    {\sum \limits_{(y,x) \in \Omega^{i,m}_n}} h^i_{n,k,y,x}
\end{equation}
\begin{equation}
    \sigma^{i,m}_{n,k} = \sqrt{
        {1 \over |\Omega^{i,m}_n|}
        {\sum \limits_{(y,x) \in \Omega^{i,m}_n}} (h^i_{n,k,y,x} - \mu^{i,m}_{n,k})^2
    }, 
\end{equation}
where $\Omega^{i,m}_n$ denotes the set of pixels in region $m$, which is $M_{misalign}$ or the other region, and $|\Omega^{i,m}_n|$ is the number of pixels in $\Omega^{i,m}_n$.
Similar to instance normalization~\cite{ulyanov2016instance}, the activation is standardized per channel.
However, \norm normalization divides the activation in channel $k$ into the activation in the misaligned region and the other region.

The rationale behind this strategy is to remove the misleading information in the misaligned regions.
Specifically, the misaligned regions in the warped clothing image match the background that is irrelevant to the clothing texture.
Performing a standardization separately on these regions leads to a removal of the background information that causes the artifacts in the final results.
In modulation, affine parameters inferred from the segmentation map modulate the standardized activation.
Due to injecting semantic information at each \norm normalization layer, the layout of the human-parsing map in the final result is preserved.

\textbf{\norm Generator.}
Fig.~\ref{fig:try-on synthesis} describes the overview of the \norm generator, which adopts the simplified architecture that discards the encoder part of an encoder-decoder network.
The generator employs a series of residual blocks (ResBlk) with upsampling layers.
Each \norm ResBlk consists of three convolutional layers and three \norm normalization layers. 
Due to the different resolutions that ResBlks operate at, we resize the inputs of the normalization layers, $\hat{S}$ and $M_{misalign}$, before injecting them into each layer.
Similarly, the input of the generator, $(I_a, P, \mathcal{W}(c, \theta))$, is resized to different resolutions.
Before each ResBlk, the resized inputs $(I_a, P, \mathcal{W}(c, \theta))$ are concatenated to the activation of the previous layer after passing through a convolution layer, and each ResBlk utilizes the concatenated inputs to refine the activation.
In this manner, the network performs the multi-scale refinement at a feature level that better preserves the clothing details than a single refinement at the pixel level.
We train the \norm generator with the conditional adversarial loss, the feature matching loss, and the perceptual loss following SPADE~\cite{park2019semantic} and pix2pixHD~\cite{wang2018high}.
Details of the model architecture, hyperparameters, and the loss function are described in the supplementary.

\section{Experiments}

\begin{figure*}[t!]
    \centering
    \includegraphics[width=\linewidth]{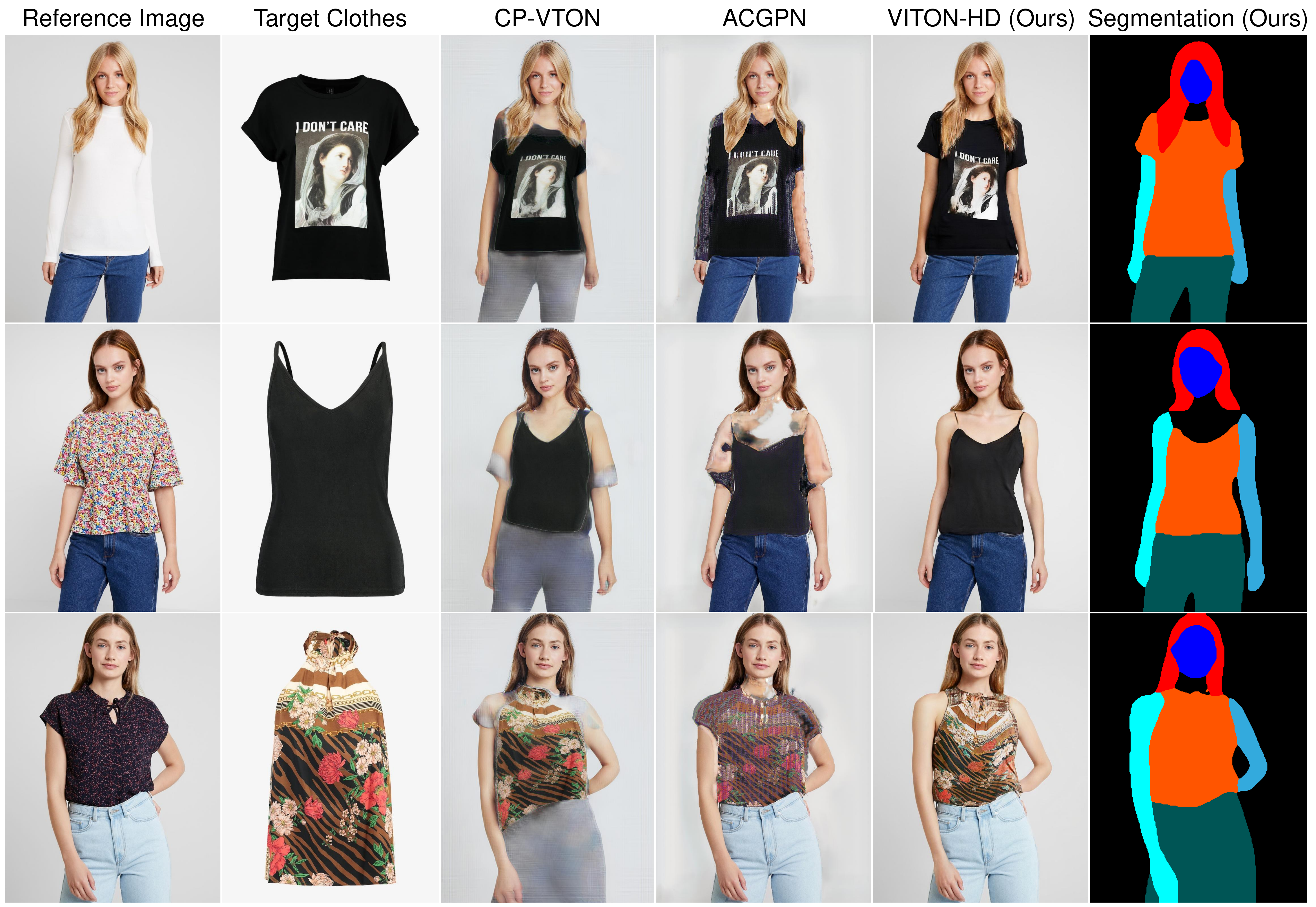}
    \vspace{-0.6cm}
    \caption{Qualitative comparison of the baselines.
    }
    \vspace{-0.3cm}
    \label{fig:experiment-baselines}
\end{figure*}

\begin{figure}[t!]
    \centering
    \includegraphics[width=1.0\linewidth]{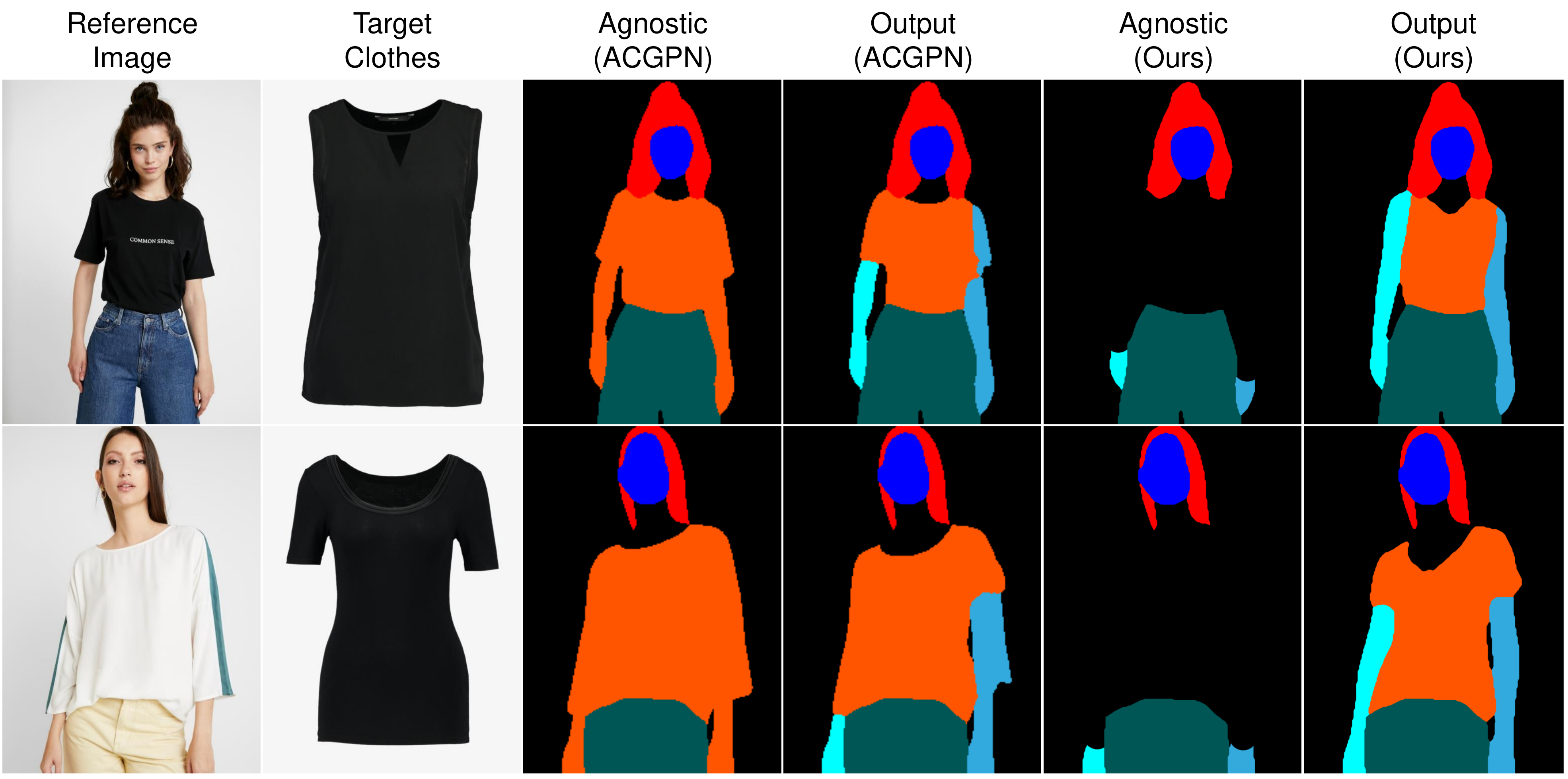}
    \vspace{-0.6cm}
    \caption{Qualitative comparison of the segmentation generator of ACGPN and \model. The clothing-agnostic segmentation map used by each model is also reported.}
    \vspace{-0.5cm}
    \label{fig:experiment-segment}
\end{figure}

\begin{figure*}[t!]
\centering
\parbox{0.64\textwidth}{
    \includegraphics[width=\linewidth]{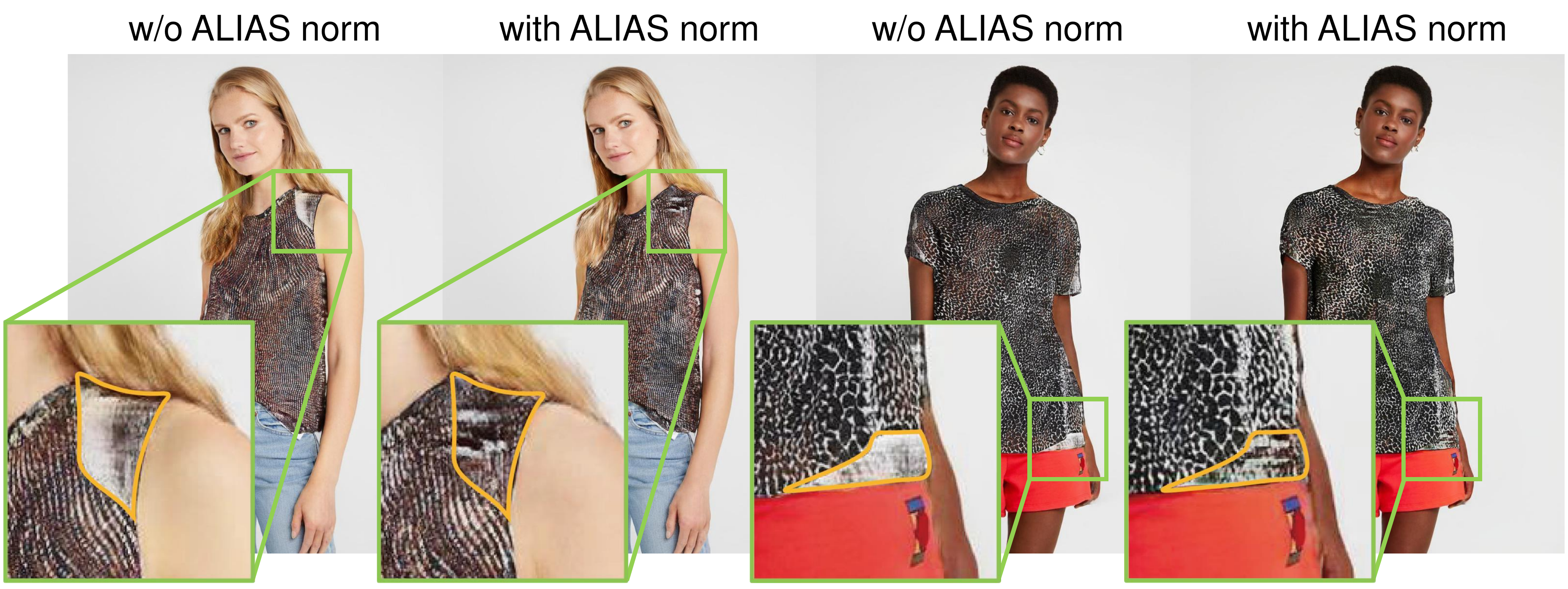}
    \vspace{-0.5cm}
    \caption{Effects of \norm normalization. The orange colored areas in the enlarged images indicate the misaligned regions.}
    \label{fig:experiment-effectiveness of alias}
}
\hfill
\centering
\parbox{0.34\textwidth}{
    \includegraphics[width=\linewidth]{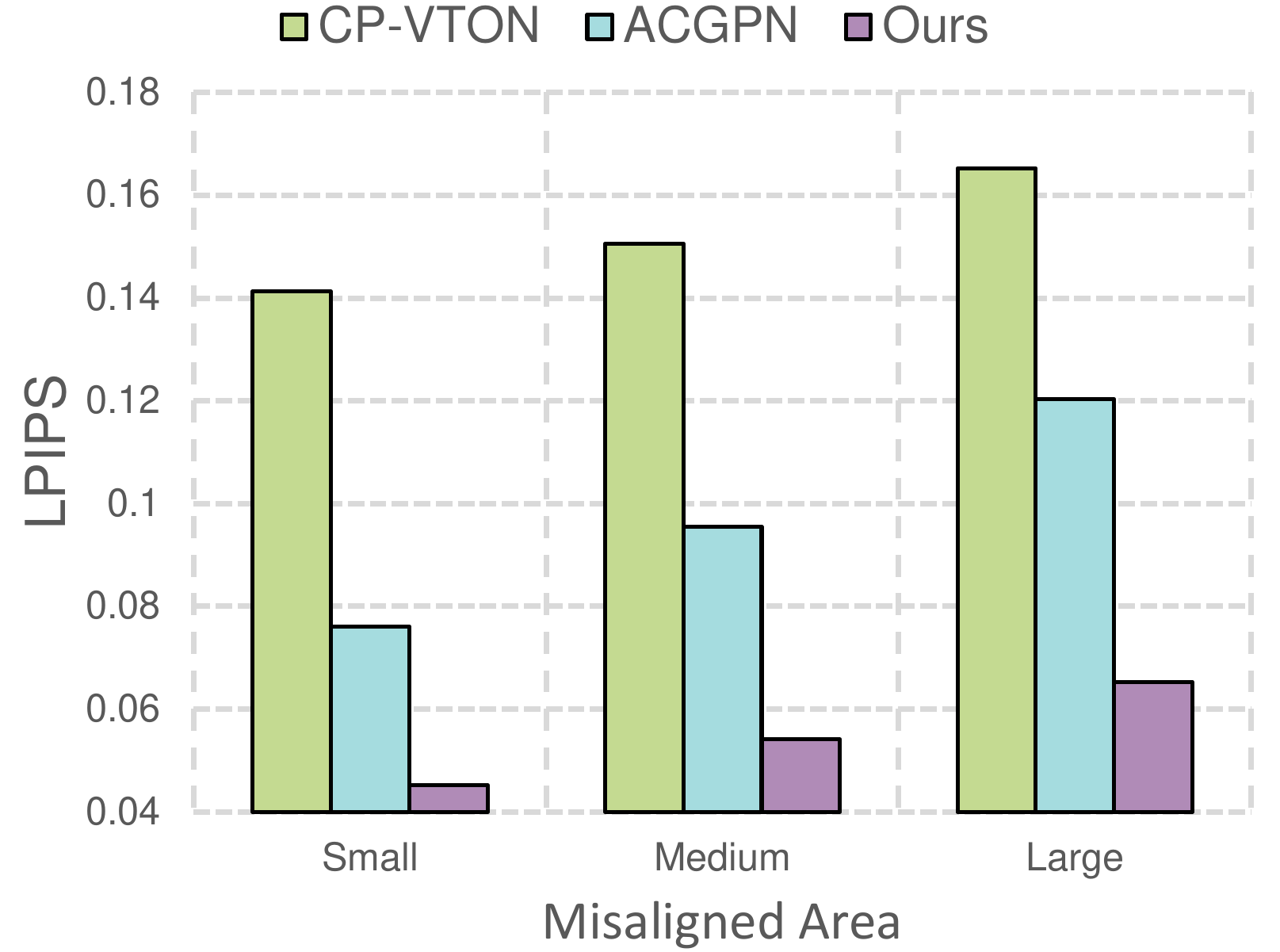}
    \vspace{-0.9cm}
    \caption{LPIPS scores according to the degree of misalignment.}
    \label{fig:experiment-misalign graph}
}
\vspace{-0.4cm}
\end{figure*}

\subsection{Experiment Setup}

\textbf{Dataset.}
We collected 1024$\times$768 virtual try-on dataset for our research purpose, since the resolution of images on the dataset provided by Han\etal~\cite{han2018viton} was low.
Specifically, we crawled 13,679 frontal-view woman and top clothing image pairs on an online shopping mall website.
The pairs are split into a training and a test set with 11,647 and 2,032 pairs, respectively.
We use the pairs of a person and a clothing image to evaluate a paired setting, and we shuffle the clothing images for an unpaired setting.
The paired setting is to reconstruct the person image with the original clothing item, and the unpaired setting is to change the clothing item on the person image with a different item.

\textbf{Training and Inference.}
With the goal of reconstructing $I$ from $(I_a, c)$, the training of each stage proceeds individually.
During the training of the geometric matching module and the \norm generator, we use $S$ instead of $\hat{S}$.
While we aim to generate 1024$\times$768 try-on images, we train the segmentation generator and the geometric matching module at 256$\times$192.
In the inference phase, after being predicted by the segmentation generator at 256$\times$192, the segmentation map is upscaled to 1024$\times$768 and passed to subsequent stages.
Similarly, the geometric matching module predicts the TPS parameters $\theta$ at 256$\times$192, and the 1024$\times$768 clothing image deformed by the parameters $\theta$ is used in the \norm generator.
We empirically found that this approach makes these two modules perform better with a lower memory cost than those trained at 1024$\times$768. Details of the model architecture and hyperparameters are described in the supplementary.

\subsection{Qualitative Analysis}
We compare \model with CP-VTON~\cite{wang2018toward} and ACGPN~\cite{yang2020towards}, whose codes are publicly available.
Following the training and inference procedure of our model, segmentation generators and geometric matching modules of the baselines are trained at 256$\times$192, and the outputs from the modules are upscaled to 1024$\times$768 during the inference.

\textbf{Comparison with Baselines.}
Fig.~\ref{fig:experiment-baselines} demonstrates that \model generates more perceptually convincing 1024$\times$768 images compared to the baselines.
Our model clearly preserves the details of the target clothes, such as the logos and the clothing textures, due to the multi-scale refinement at a feature level.
In addition, regardless of what clothes the person is wearing in the reference image, our model synthesizes the body shape naturally.
As shown in Fig.~\ref{fig:experiment-segment}, the shape of the original clothing item remains in the synthetic segmentation map generated by ACGPN.
On the other hand, the segmentation generator in \model successfully predicts the segmentation map regardless of the original clothing item, due to our newly proposed clothing-agnostic person representation.
Although our model surpasses the baselines qualitatively, there are a few limitations to \model, which are reported in the supplementary with the additional qualitative results.

\textbf{Effectiveness of the \norm Normalization.}
We study the effectiveness of \norm normalization by comparing our model to VITON-HD*, where the standardization in \norm normalization is replaced by channel-wise standardization, as in the original instance normalization~\cite{ulyanov2016instance}.
Fig.~\ref{fig:experiment-effectiveness of alias} shows that \norm normalization has the capability to fill the misaligned areas with the target clothing texture by removing the misleading information.
On the other hand, without utilizing \norm normalization, the artifacts are produced in the misaligned areas, because the background information in the warped clothing image is not removed as described in Section~\ref{sec:try-on synthesis}.
\norm normalization, however, can handle the misaligned regions properly.

\subsection{Quantitative Analysis}
We perform the quantitative experiments in both a paired and an unpaired settings, in which a person wears the original clothes or the new clothes, respectively.
We evaluate our method using three metrics widely used in virtual try-on.
The structural similarity (SSIM)~\cite{wang2004image} and the learned perceptual image patch similarity (LPIPS)~\cite{zhang2018unreasonable} are used in the paired setting, and the frechet inception distance (FID)~\cite{heusel2017gans} score is adopted in the unpaired setting.
The inception score~\cite{salimans2016improved} is not included in the experiments, since it cannot reflect whether the details of the clothing image are maintained~\cite{han2018viton}.
The input of the each model contains different amount of information that offers advantages in reconstructing the segmentation maps, thus we use the segmentation maps from the test set instead of the synthetic segmentation maps in the paired setting.

\textbf{Comparison across Different Resolutions.}
We compare the baselines quantitatively across different resolutions (256$\times$192, 512$\times$384, and 1024$\times$768) as shown in Table~\ref{Table:resolution comparison}.
Our model outperforms the baselines for SSIM and LPIPS across all resolutions.
For FID score, our model significantly surpasses CP-VTON, regardless of the resolutions.
The FID score in ACGPN is slightly lower than that of our model at the 256$\times$192 resolution.
However, at the 1024$\times$768 resolution, our model achieves a lower FID score than ACGPN with a large margin.
The results indicate that the baselines cannot handle 1024$\times$768 images, while our model is trained in a stable manner, even at a high resolution.
This may be due to the limited capability of the U-Net architecture employed in the baseline models.

\textbf{Comparison According to the Degree of Misalignment.}
To verify the ability of filling the misaligned areas with the clothing texture, we perform experiments in the paired setting according to the degree of the misalignment.
According to the number of pixels in the misaligned areas, we divide the test dataset in three types: small, medium, and large.
For a fair comparison, each model uses the same segmentation maps and the same warped clothes as inputs to match the misaligned regions.
We evaluate LPIPS to measure the semantic distances between the reference images and the reconstructed images.
As shown in Fig.~\ref{fig:experiment-misalign graph}, the wider the misaligned areas, the worse the performance of models, which means that the misalignment hinders the models from generating photo-realistic virtual try-on images.
Compared to the baselines, our model consistently performs better, and the performance of our model decreases less as the degree of misalignment increases.

\section{Conclusions}
We propose the \model that synthesizes photo-realistic 1024$\times$768 virtual try-on images.
The proposed \norm normalization can properly handle the misaligned areas and propagate the semantic information throughout the \norm generator, which preserves the details of the clothes via the multi-scale refinement.
Qualitative and quantitative experiments demonstrate that \model surpasses existing virtual try-on methods with a large margin.

\vspace{0.2cm}
\noindent\textbf{Acknowledgments.}
This work was supported by the National Research Foundation of Korea (NRF) grant funded by the Korean government (MSIT) (No. NRF-2019R1A2C4070420) and Seoul R\&BD Program (CD200024) through the Seoul Business Agency (SBA) funded by the Seoul Metropolitan Government.

{
\small
\bibliographystyle{ieee_fullname}
\bibliography{reference}

\begin{thebibliography}{10}\itemsep=-1pt

\bibitem{anokhin2020high}
Ivan Anokhin, Pavel Solovev, Denis Korzhenkov, Alexey Kharlamov, Taras
  Khakhulin, Aleksei Silvestrov, Sergey Nikolenko, Victor Lempitsky, and Gleb
  Sterkin.
\newblock High-resolution daytime translation without domain labels.
\newblock In {\em Proc. of the IEEE conference on computer vision and pattern
  recognition (CVPR)}, pages 7488--7497, 2020.

\bibitem{brock2018large}
Andrew Brock, Jeff Donahue, and Karen Simonyan.
\newblock Large scale gan training for high fidelity natural image synthesis.
\newblock In {\em Proc. the International Conference on Learning
  Representations (ICLR)}, 2018.

\bibitem{cao2019openpose}
Z Cao, T Simon, SE Wei, YA Sheikh, et~al.
\newblock Openpose: Realtime multi-person 2d pose estimation using part
  affinity fields.
\newblock {\em The IEEE Transactions on Pattern Analysis and Machine
  Intelligence (TPAMI)}, 2019.

\bibitem{de2017modulating}
Harm De~Vries, Florian Strub, J{\'e}r{\'e}mie Mary, Hugo Larochelle, Olivier
  Pietquin, and Aaron~C Courville.
\newblock Modulating early visual processing by language.
\newblock In {\em Proc. the Advances in Neural Information Processing Systems
  (NeurIPS)}, pages 6594--6604, 2017.

\bibitem{dong2019towards}
Haoye Dong, Xiaodan Liang, Xiaohui Shen, Bochao Wang, Hanjiang Lai, Jia Zhu,
  Zhiting Hu, and Jian Yin.
\newblock Towards multi-pose guided virtual try-on network.
\newblock In {\em Proc. of the IEEE international conference on computer vision
  (ICCV)}, pages 9026--9035, 2019.

\bibitem{dong2019fw}
Haoye Dong, Xiaodan Liang, Xiaohui Shen, Bowen Wu, Bing-Cheng Chen, and Jian
  Yin.
\newblock Fw-gan: Flow-navigated warping gan for video virtual try-on.
\newblock In {\em Proc. of the IEEE international conference on computer vision
  (ICCV)}, pages 1161--1170, 2019.

\bibitem{gong2018instance}
Ke Gong, Xiaodan Liang, Yicheng Li, Yimin Chen, Ming Yang, and Liang Lin.
\newblock Instance-level human parsing via part grouping network.
\newblock In {\em Proc. of the European Conference on Computer Vision (ECCV)},
  pages 770--785, 2018.

\bibitem{guan2012drape}
Peng Guan, Loretta Reiss, David~A Hirshberg, Alexander Weiss, and Michael~J
  Black.
\newblock Drape: Dressing any person.
\newblock {\em ACM Transactions on Graphics (TOG)}, 31(4):1--10, 2012.

\bibitem{han2019clothflow}
Xintong Han, Xiaojun Hu, Weilin Huang, and Matthew~R Scott.
\newblock Clothflow: A flow-based model for clothed person generation.
\newblock In {\em Proc. of the IEEE international conference on computer vision
  (ICCV)}, pages 10471--10480, 2019.

\bibitem{han2018viton}
Xintong Han, Zuxuan Wu, Zhe Wu, Ruichi Yu, and Larry~S Davis.
\newblock Viton: An image-based virtual try-on network.
\newblock In {\em Proc. of the IEEE conference on computer vision and pattern
  recognition (CVPR)}, pages 7543--7552, 2018.

\bibitem{heusel2017gans}
Martin Heusel, Hubert Ramsauer, Thomas Unterthiner, Bernhard Nessler, and Sepp
  Hochreiter.
\newblock Gans trained by a two time-scale update rule converge to a local nash
  equilibrium.
\newblock In {\em Proc. the Advances in Neural Information Processing Systems
  (NeurIPS)}, pages 6629--6640, 2017.

\bibitem{huang2017arbitrary}
Xun Huang and Serge Belongie.
\newblock Arbitrary style transfer in real-time with adaptive instance
  normalization.
\newblock In {\em Proc. of the IEEE international conference on computer vision
  (ICCV)}, pages 1501--1510, 2017.

\bibitem{ioffe2015batch}
Sergey Ioffe and Christian Szegedy.
\newblock Batch normalization: Accelerating deep network training by reducing
  internal covariate shift.
\newblock In {\em Proc. the International Conference on Machine Learning
  (ICML)}, pages 448--456, 2015.

\bibitem{isola2017image}
Phillip Isola, Jun-Yan Zhu, Tinghui Zhou, and Alexei~A Efros.
\newblock Image-to-image translation with conditional adversarial networks.
\newblock In {\em Proc. of the IEEE conference on computer vision and pattern
  recognition (CVPR)}, pages 1125--1134, 2017.

\bibitem{jetchev2017conditional}
Nikolay Jetchev and Urs Bergmann.
\newblock The conditional analogy gan: Swapping fashion articles on people
  images.
\newblock In {\em Proc. of the IEEE international conference on computer vision
  workshop (ICCVW)}, pages 2287--2292, 2017.

\bibitem{kingma2014adam}
Diederik~P Kingma and Jimmy Ba.
\newblock Adam: A method for stochastic optimization.
\newblock {\em arXiv:1412.6980}, 2014.

\bibitem{mao2017least}
Xudong Mao, Qing Li, Haoran Xie, Raymond~YK Lau, Zhen Wang, and Stephen
  Paul~Smolley.
\newblock Least squares generative adversarial networks.
\newblock In {\em Proc. of the IEEE international conference on computer vision
  (ICCV)}, pages 2794--2802, 2017.

\bibitem{miyato2018spectral}
Takeru Miyato, Toshiki Kataoka, Masanori Koyama, and Yuichi Yoshida.
\newblock Spectral normalization for generative adversarial networks.
\newblock In {\em Proc. the International Conference on Learning
  Representations (ICLR)}, 2018.

\bibitem{odena2017conditional}
Augustus Odena, Christopher Olah, and Jonathon Shlens.
\newblock Conditional image synthesis with auxiliary classifier gans.
\newblock In {\em Proc. the International Conference on Machine Learning
  (ICML)}, pages 2642--2651, 2017.

\bibitem{park2019semantic}
Taesung Park, Ming-Yu Liu, Ting-Chun Wang, and Jun-Yan Zhu.
\newblock Semantic image synthesis with spatially-adaptive normalization.
\newblock In {\em Proc. of the IEEE conference on computer vision and pattern
  recognition (CVPR)}, pages 2337--2346, 2019.

\bibitem{park2020swapping}
Taesung Park, Jun-Yan Zhu, Oliver Wang, Jingwan Lu, Eli Shechtman, Alexei~A.
  Efros, and Richard Zhang.
\newblock Swapping autoencoder for deep image manipulation.
\newblock In {\em Proc. the Advances in Neural Information Processing Systems
  (NeurIPS)}, 2020.

\bibitem{patel2020tailornet}
Chaitanya Patel, Zhouyingcheng Liao, and Gerard Pons-Moll.
\newblock Tailornet: Predicting clothing in 3d as a function of human pose,
  shape and garment style.
\newblock In {\em Proc. of the IEEE conference on computer vision and pattern
  recognition (CVPR)}, pages 7365--7375, 2020.

\bibitem{pons2017clothcap}
Gerard Pons-Moll, Sergi Pujades, Sonny Hu, and Michael~J Black.
\newblock Clothcap: Seamless 4d clothing capture and retargeting.
\newblock {\em ACM Transactions on Graphics (TOG)}, 36(4):1--15, 2017.

\bibitem{reed2016generative}
Scott Reed, Zeynep Akata, Xinchen Yan, Lajanugen Logeswaran, Bernt Schiele, and
  Honglak Lee.
\newblock Generative adversarial text to image synthesis.
\newblock In {\em Proc. the International Conference on Machine Learning
  (ICML)}, pages 1060--1069, 2016.

\bibitem{ronneberger2015u}
Olaf Ronneberger, Philipp Fischer, and Thomas Brox.
\newblock U-net: Convolutional networks for biomedical image segmentation.
\newblock In {\em International Conference on Medical Image Computing and
  Computer Assisted Intervention}, pages 234--241, 2015.

\bibitem{salimans2016improved}
Tim Salimans, Ian Goodfellow, Wojciech Zaremba, Vicki Cheung, Alec Radford, and
  Xi Chen.
\newblock Improved techniques for training gans.
\newblock In {\em Proc. the Advances in Neural Information Processing Systems
  (NeurIPS)}, pages 2234--2242, 2016.

\bibitem{sekine2014virtual}
Masahiro Sekine, Kaoru Sugita, Frank Perbet, Bj{\"o}rn Stenger, and Masashi
  Nishiyama.
\newblock Virtual fitting by single-shot body shape estimation.
\newblock In {\em International Conference on 3D Body Scanning Technologies},
  pages 406--413, 2014.

\bibitem{shen2017learning}
Wei Shen and Rujie Liu.
\newblock Learning residual images for face attribute manipulation.
\newblock In {\em Proc. of the IEEE conference on computer vision and pattern
  recognition (CVPR)}, pages 4030--4038, 2017.

\bibitem{simonyan2014very}
Karen Simonyan and Andrew Zisserman.
\newblock Very deep convolutional networks for large-scale image recognition.
\newblock {\em arXiv preprint arXiv:1409.1556}, 2014.

\bibitem{ulyanov2016instance}
Dmitry Ulyanov, Andrea Vedaldi, and Victor Lempitsky.
\newblock Instance normalization: The missing ingredient for fast stylization.
\newblock {\em arXiv preprint arXiv:1607.08022}, 2016.

\bibitem{wang2018toward}
Bochao Wang, Huabin Zheng, Xiaodan Liang, Yimin Chen, Liang Lin, and Meng Yang.
\newblock Toward characteristic-preserving image-based virtual try-on network.
\newblock In {\em Proc. of the European Conference on Computer Vision (ECCV)},
  pages 589--604, 2018.

\bibitem{wang2018high}
Ting-Chun Wang, Ming-Yu Liu, Jun-Yan Zhu, Andrew Tao, Jan Kautz, and Bryan
  Catanzaro.
\newblock High-resolution image synthesis and semantic manipulation with
  conditional gans.
\newblock In {\em Proc. of the IEEE conference on computer vision and pattern
  recognition (CVPR)}, pages 8798--8807, 2018.

\bibitem{wang2004image}
Zhou Wang, Alan~C Bovik, Hamid~R Sheikh, and Eero~P Simoncelli.
\newblock Image quality assessment: from error visibility to structural
  similarity.
\newblock {\em IEEE Transactions on Image Processing}, 13(4):600--612, 2004.

\bibitem{xu2018attngan}
Tao Xu, Pengchuan Zhang, Qiuyuan Huang, Han Zhang, Zhe Gan, Xiaolei Huang, and
  Xiaodong He.
\newblock Attngan: Fine-grained text to image generation with attentional
  generative adversarial networks.
\newblock In {\em Proc. of the IEEE conference on computer vision and pattern
  recognition (CVPR)}, pages 1316--1324, 2018.

\bibitem{yang2020towards}
Han Yang, Ruimao Zhang, Xiaobao Guo, Wei Liu, Wangmeng Zuo, and Ping Luo.
\newblock Towards photo-realistic virtual try-on by adaptively
  generating-preserving image content.
\newblock In {\em Proc. of the IEEE conference on computer vision and pattern
  recognition (CVPR)}, pages 7850--7859, 2020.

\bibitem{yu2019vtnfp}
Ruiyun Yu, Xiaoqi Wang, and Xiaohui Xie.
\newblock Vtnfp: An image-based virtual try-on network with body and clothing
  feature preservation.
\newblock In {\em Proc. of the IEEE international conference on computer vision
  (ICCV)}, pages 10511--10520, 2019.

\bibitem{zhang2019self}
Han Zhang, Ian Goodfellow, Dimitris Metaxas, and Augustus Odena.
\newblock Self-attention generative adversarial networks.
\newblock In {\em Proc. the International Conference on Machine Learning
  (ICML)}, pages 7354--7363, 2019.

\bibitem{zhang2018unreasonable}
Richard Zhang, Phillip Isola, Alexei~A Efros, Eli Shechtman, and Oliver Wang.
\newblock The unreasonable effectiveness of deep features as a perceptual
  metric.
\newblock In {\em Proc. of the IEEE conference on computer vision and pattern
  recognition (CVPR)}, 2018.

\bibitem{zhu2020sean}
Peihao Zhu, Rameen Abdal, Yipeng Qin, and Peter Wonka.
\newblock Sean: Image synthesis with semantic region-adaptive normalization.
\newblock In {\em Proc. of the IEEE conference on computer vision and pattern
  recognition (CVPR)}, pages 5104--5113, 2020.

\end{thebibliography}
}

\newpage
\twocolumn[
\begin{center}
    \vspace*{1.1cm}
    \Large{\bf{Supplementary Material}}
    \vspace*{1.7cm}
\end{center}]

\section*{A. Implementation Details}

\subsection*{A.1. Pre-processing Details}
This section introduces the details of generating our clothing-agnostic person representation.
To remove the dependency on the clothing item originally worn by a person, regions that can provide any original clothing information, such as the arms that hint at the sleeve length, should be eliminated.
Therefore, when generating a clothing-agnostic image $I_a$, we remove the arms from the reference image $I$.
For the same reason, legs should be removed if the pants are the target clothing items.
We mask the regions with a gray color, so that the masked pixels of the normalized image would have a value of 0.
We add padding to the masks to thoroughly remove these regions, and the width of the padding is empirically determined.

\subsection*{A.2. Model Architectures}
This section introduces the architectures of the segmentation generator, the geometric matching module, and \norm generator in detail.

\begin{figure}[b!]
    \centering
    \includegraphics[width=0.8\linewidth]{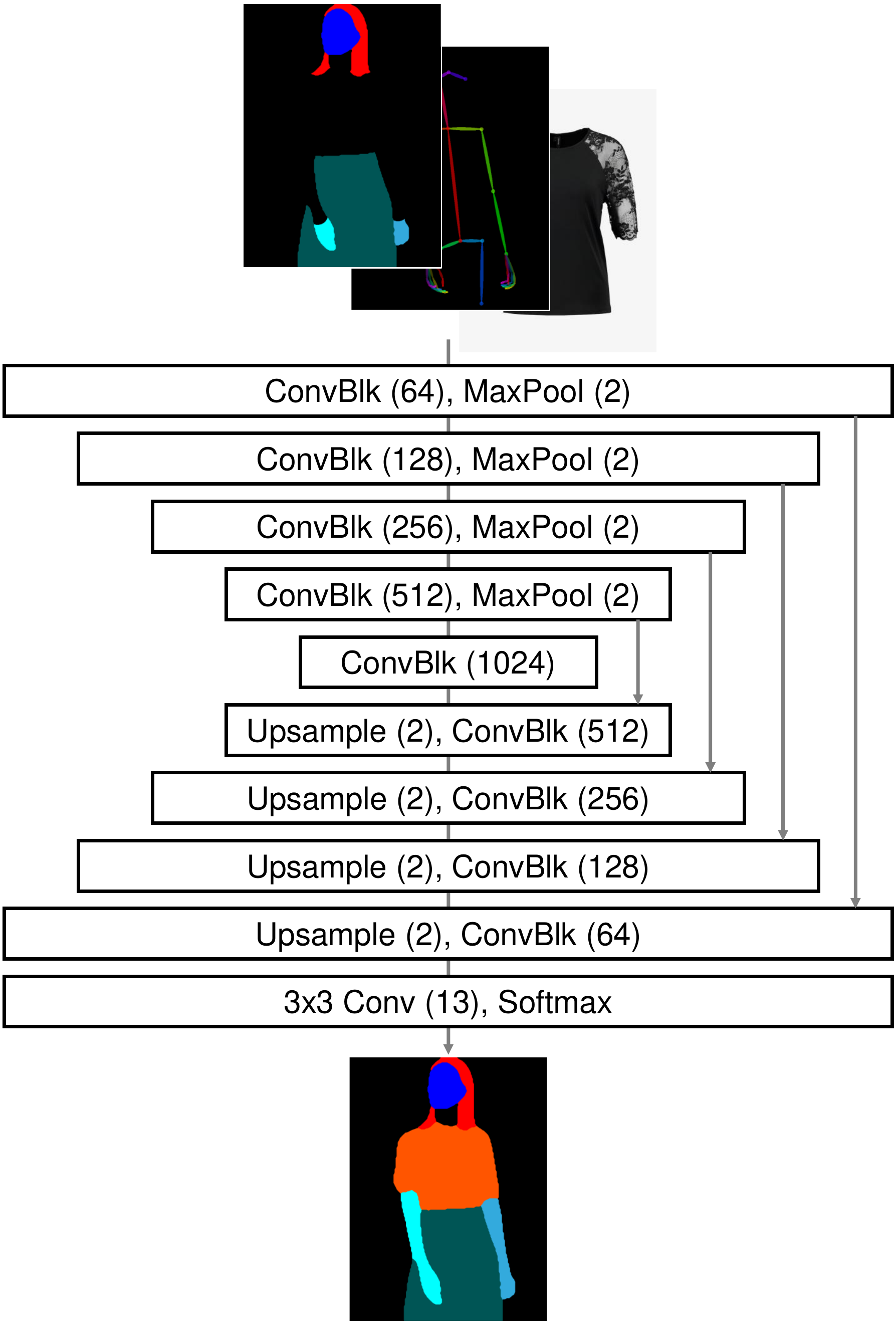}
    \caption{Segmentation Generator. 
    $k\times{k}$ Conv ($x$) denotes a convolutional layer where the kernel size is $k$ and the output channel is $x$.
    Also, ConvBlk ($x$) denotes a block, which consists of two series of 3$\times$3 convolutional layer, instance normalization, and ReLU activation.}
    \label{fig:supp-segmentation generator}
\end{figure}

\textbf{Segmentation Generator.}
The segmentation generator has the structure of U-Net~\cite{ronneberger2015u}, which consists of convolutional layers, downsampling layers, and upsampling layers.
Two multi-scale discriminators~\cite{wang2018high} are employed for the conditional adversarial loss.
The details of the segmentation generator architecture are shown in Fig.~\ref{fig:supp-segmentation generator}.

\begin{figure}[t!]
    \centering
    \includegraphics[width=0.8\linewidth]{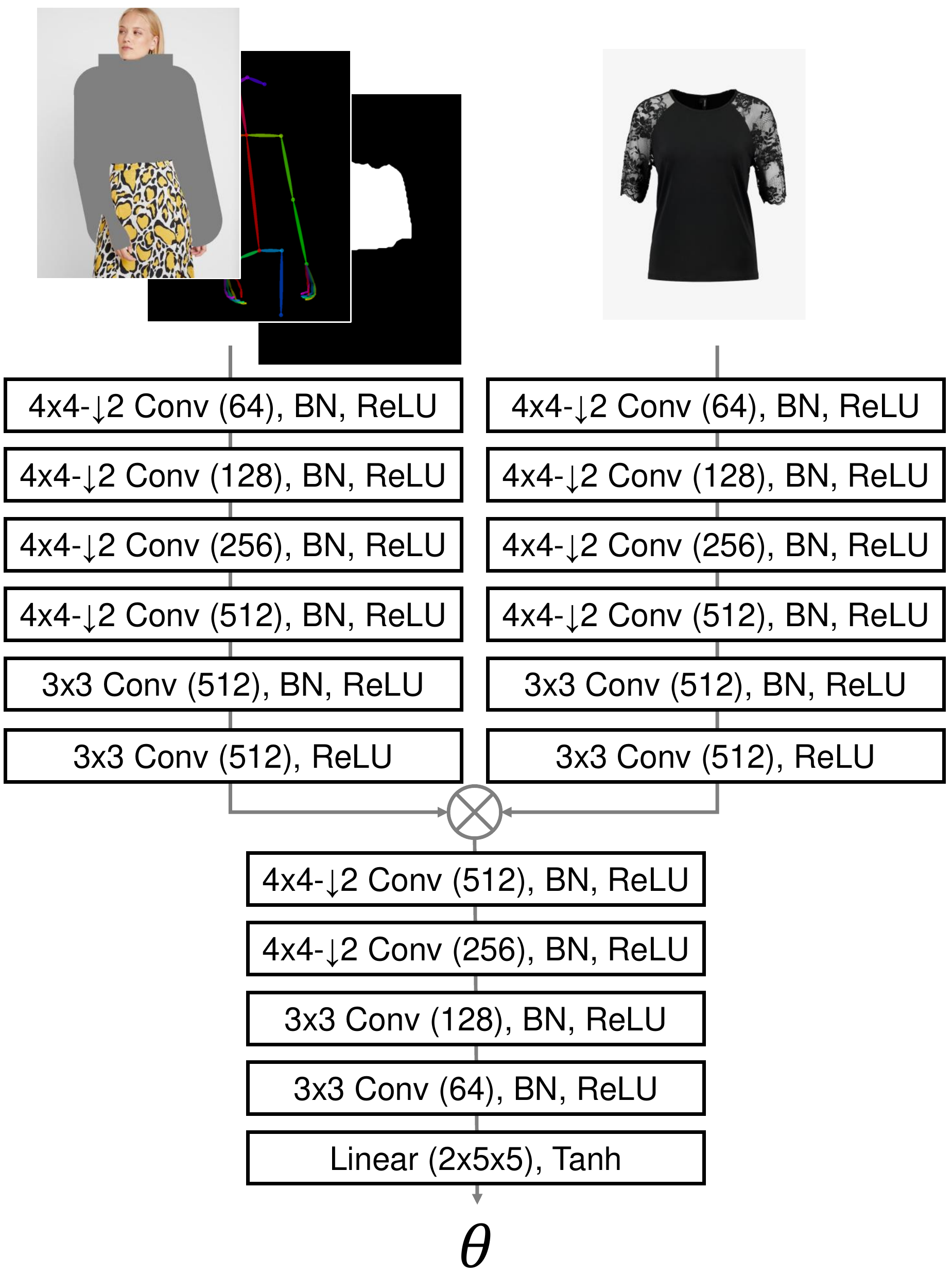}
    \caption{Geometric Matching Module.
    $k\times{k}\downarrow$2 Conv ($x$) denotes a convolutional layer where the kernel size is $k$, the stride is 2, and the output channel is $x$.}
    \label{fig:supp-geometric matching module}
\end{figure}

\textbf{Geometric Matching Module.}
The geometric matching module consists of two feature extractors and a regression network.
A correlation matrix is calculated from the two extracted features, and the regression network predicts the TPS parameter $\theta$ with the correlation matrix.
The feature extractor is composed of a series of convolutional layers, and the regression network consists of a series of convolutional layers followed by a fully connected layer. 
The details are shown in Fig.~\ref{fig:supp-geometric matching module}.

\begin{figure}[t!]
    \centering
    \includegraphics[width=0.8\linewidth]{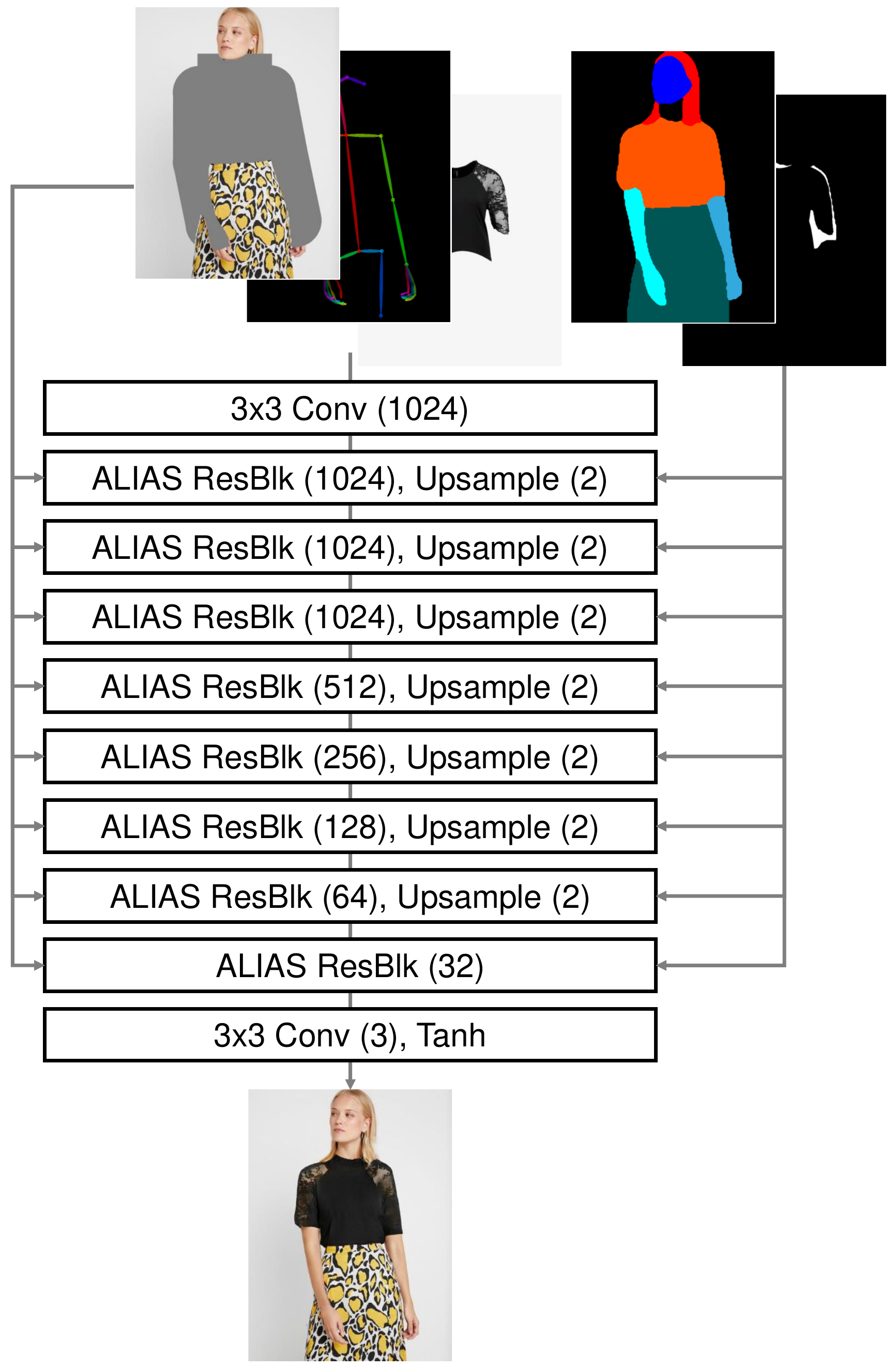}
    \caption{\norm Generator. The segmentation map $S$ and the misalignment mask $M_{misalign}$ are passed to the generator through the proposed \norm ResBlks.}
    \label{fig:supp-alias generator}
\end{figure}

\textbf{\norm Generator.}
The architecture of the \norm generator consists of a series of \norm ResBlks with nearest-neighbor upsampling layers.
We employ two multi-scale discriminators with instance normalization.
Spectral normalization~\cite{miyato2018spectral} is applied to all the convolutional layers.
Note that we separately standardize the activation based on the misalignment mask $M_{misalign}$ only in the first five \norm ResBlks.
The details of the \norm generator architecture is shown in Fig.~\ref{fig:supp-alias generator}.

\subsection*{A.3. Training Details}
This section introduces the losses and the hyperparameters for the segmentation generator, the geometric matching module, and the \norm generator.

\textbf{Segmentation Generator.}
The segmentation generator $G_S$ uses the clothing-agnostic segmentation map $S_a$, the pose map $P$, and the clothing item $c$ as inputs ($\hat{S} = G_S(S_a, P, c)$) to predict the segmentation map $\hat{S}$ of the person in the reference image wearing the target clothing item.
The segmentation generator is trained with the cross-entropy loss $\mathcal{L}_{CE}$ and the conditional adversarial loss $\mathcal{L}_{cGAN}$, which is LSGAN loss~\cite{mao2017least}. 
The full loss $\mathcal{L}_S$ for the segmentation generator are written as
\begin{equation}
    \mathcal{L}_S = \mathcal{L}_{cGAN} + \lambda_{CE} \mathcal{L}_{CE}
    \label{eq:sup_segloss}
\end{equation}
\begin{equation}
    \mathcal{L}_{CE} = -{1\over{HW}}
    {\sum\limits_{{k}\in{C}, {y}\in{H}, {x}\in{W}}{S_{k, y, x}\log(\hat{S}_{k, y, x})}}
    \label{eq:segCEloss}
\end{equation}
\begin{equation}
    \begin{split}
        \mathcal{L}_{cGAN} &= \mathbb{E}_{(X,S)}[\log(D(X, S))] \\
        &+ \mathbb{E}_{X}[1 - \log(D(X, \hat{S}))],
    \end{split}
    \label{eq:segcGANloss}
\end{equation}
where $\lambda_{CE}$ is the hyperparameter for the cross-entropy loss.
In the experiment, $\lambda_{CE}$ is set to 10.
In Eq.~\eqref{eq:segCEloss}, $S_{yxk}$ and $\hat{S}_{yxk}$ indicate the pixel values of the segmentation map of the reference image $S$ and $\hat{S}$ corresponding to the coordinates $(x,y)$ in channel $k$. 
The symbols $H$, $W$ and $C$ indicate the height, width, and the number of channels of $S$.
In Eq.~\eqref{eq:segcGANloss}, the symbol $X$ indicates the inputs of the generator ($S_a,P,c$), and $D$ denotes the discriminator.

The learning rate of the generator and the discriminator is 0.0004.
We adopt the Adam optimizer with $\beta_1 = 0.5$ and $\beta_2 = 0.999$.
We train the segmentation generator for 200,000 iterations with the batch size of 8.

\textbf{Geometric Matching Module.}
The inputs of the geometric matching module are $c$, $P$, clothing-agnostic image $I_a$, and $\hat{S_c}$, which is the clothing area of $\hat{S}$.
The output is the TPS transformation parameters $\theta$.
The overall objective function is written as
\begin{equation}
    \mathcal{L}_{warp} = ||I_{c} - \mathcal{W}(c, \theta)||_{1,1} + \lambda_{const} \mathcal{L}_{const}
\end{equation}
\begin{equation}
    \begin{split}
        \mathcal{L}_{const} = \sum\limits_{p\in{\mathbf{P}}}|\,(|\hspace{0.5mm}||pp_0||_2 - ||pp_1||_2|+|\hspace{0.5mm}||pp_2||_2-||pp_3||_2|) \\
        + (|\mathcal{S}(p, p_0) - \mathcal{S}(p, p_1)| + |\mathcal{S}(p, p_2) - \mathcal{S}(p, p_3)|),
    \end{split}
    \label{eq:secondorderloss}
\end{equation}
where $\mathcal{W}$ is the function that deforms $c$ using $\theta$, and $I_c$ is the clothing item extracted from the reference image $I$.
$\mathcal{L}_{const}$ is a second-order difference constraint~\cite{yang2020towards}, and $\lambda_{const}$ is the hyperparameter for $\mathcal{L}_{const}$.
In the experiment, we set $\lambda_{const}$ to 0.04.
In Eq.~\eqref{eq:secondorderloss}, the symbol $p$ indicates a sampled TPS control point from the entire control points set $\mathbf{P}$, and $p_0$, $p_1$, $p_2$, and $p_3$ are top, bottom, left and right point of $p$, respectively.
The function $\mathcal{S}(p, p_i)$ denotes the slope between $p$ and $p_i$.

The learning rate of the geometric matching module is 0.0002.
We adopt the Adam optimizer with $\beta_1 = 0.5$ and $\beta_2 = 0.999$.
We train the geometric matching module for 50,000 iterations with the batch size of 8. 

\textbf{\norm Generator.}
The loss function of \norm generator follows those of SPADE~\cite{park2019semantic} and pix2pixHD~\cite{wang2018high}, as it contains the conditional adversarial loss $\mathcal{L}_{cGAN}$, the feature matching loss $\mathcal{L}_{FM}$, and the perceptual loss $\mathcal{L}_{percept}$.
Let $D_I$ be the discriminator, $I$ and $c$ be the given reference and target clothing images, and $\hat{I}$ be the synthetic image generated by the generator.
$S_{div}$ is the modified version of the segmentation map $S$.
The full loss $\mathcal{L}_I$ of our generator is written as
\begin{equation}
    \mathcal{L}_{I} = \mathcal{L}_{cGAN} + \lambda_{FM} \mathcal{L}_{FM}
    + \lambda_{percept} \mathcal{L}_{percept}
    \label{eq:aliasloss}
\end{equation}
\begin{equation}
    \begin{split}
        \mathcal{L}_{cGAN} &= \mathbb{E}_I[\log(D_I(S_{div}, I))] \\
        &+ \mathbb{E}_{(I, c)}[1 - \log(D_I(S_{div}, \hat{I}))]
    \end{split}
    \label{eq:aliascGANloss}
\end{equation}
\begin{equation}
        \mathcal{L}_{FM} = \mathbb{E}_{(I, c)}{\sum\limits_{i=1}^T}{1 \over K_i}
        [||D_I^{(i)}(S_{div}, I) - D_I^{(i)}(S_{div}, \hat{I})||_{1,1}]
\end{equation}
\begin{equation}
        \mathcal{L}_{percept} = \mathbb{E}_{(I, c)}{\sum\limits_{i=1}^V}{1 \over R_i}
        [||F^{(i)}(I) - F^{(i)}(\hat{I})||_{1,1}],
\end{equation}
where $\lambda_{FM}$ and $\lambda_{percept}$ are hyperparameters.
In the experiment, both $\lambda_{FM}$ and $\lambda_{percept}$ are set to 10.
$T$ is the number of layers in $D_I$, and $D_I^{(i)}$ and $K_i$ are the activation and the number of elements in the $i$-th layer of $D_I$, respectively.
Similarly, $V$ is the number of layers used in the VGG network $F$~\cite{simonyan2014very}, and $F^{(i)}$ and $R_i$ are the activation and the number of elements in the $i$-th layer of $F$, respectively.
We replace the standard adversarial loss with the Hinge loss~\cite{zhang2019self}.

The learning rate of the generator and the discriminator is 0.0001 and 0.0004, respectively.
We adopt the Adam optimizer~\cite{kingma2014adam} with $\beta_1 = 0$ and $\beta_2 = 0.9$.
We train the \norm generator for 200,000 iterations with the batch size of 4.

\begin{figure}[t!]
    \centering
    \includegraphics[width=\linewidth]{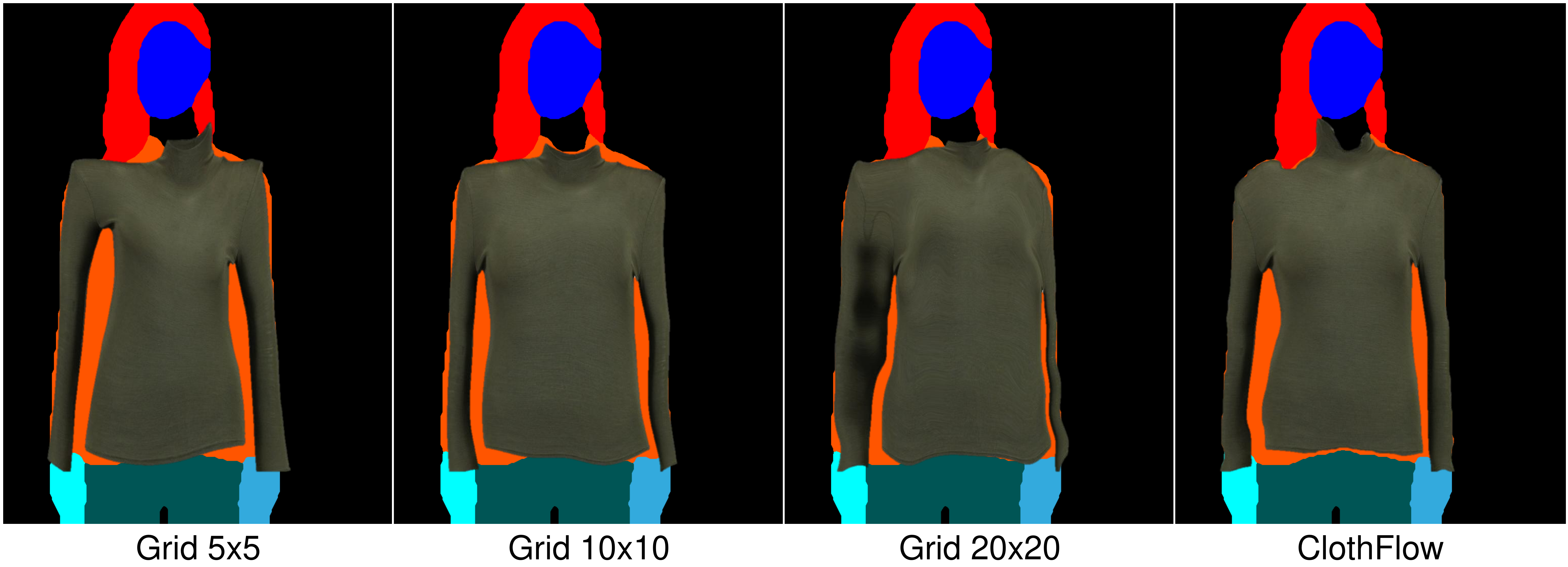}
    \vspace{-0.7cm}
    \caption{Qualitative comparisons of TPS transformation with various grid numbers and the flow estimation from ClothFlow.}
    \label{clothflow}
\end{figure}

\begin{table}[t!]
    \centering
    \begin{tabular}{c|cc|c}
        \toprule
        Method & Warp-SSIM$_{\uparrow}$ & MACs$_{\downarrow}$ & Mask-SSIM$_{\uparrow}$ \\
        \midrule
        ClothFlow & \textbf{0.841}$^{\star}$ & 8.13G & 0.803$^{\star}$ \\
        VITON-HD  & 0.782 & \textbf{4.47G} & \textbf{0.852} \\
        \bottomrule
    \end{tabular}
    \vspace{-0.1cm}
    \caption{$\star$ denotes a score taken from the ClothFlow paper, and we train VITON-HD in the same setting (\eg, dataset and resolution). We compute MACs of their warping modules at 256$\times$192.}
    \label{Table:performance}
    \vspace{-0.4cm}
\end{table}

\section*{B. Additional Experiments}

\subsection*{B.1. Comparison with ClothFlow}
To demonstrate that the optical flow estimation does not solve the misalignment completely, we re-implement the flow estimation module of ClothFlow~\cite{han2019clothflow} based on the original paper.
Fig.~\ref{clothflow} shows that the misalignment still occurs, although both TPS with a higher grid number (e.g., a 10$\times$10 or 20$\times$20 grid) and the flow estimation module of ClothFlow can reduce the misaligned regions.
The reason is that the regularization to avoid the artifacts (\eg, TV loss) prevents the warped clothes from fitting perfectly into the target region.
In addition, we evaluate the accuracy and the computational cost of warping modules in VITON-HD and ClothFlow with Warp-SSIM~\cite{han2019clothflow} and MACs, respectively.
We also measure how well the models reconstruct the clothing using Mask-SSIM~\cite{han2019clothflow}.
Table~\ref{Table:performance} shows that the ClothFlow warping module has the better accuracy than ours, whereas the higher Mask-SSIM in VITON-HD proves that ALIAS normalization is more effective at solving the misalignment problem than the improved warping method.
We found that the ClothFlow warping module needs a huge computational cost (MACs: 130.03G) at 1024$\times$768, but the cost could be reduced when predicting the optical flow map at 256$\times$192.
Table~\ref{Table:performance} demonstrates that the ClothFlow warping module still needs more computational cost than ours, yet it is a viable option to combine the flow estimation module with ALIAS generator.

\begin{figure}[b!]
    \centering
    \vspace{0.5cm}
    \includegraphics[width=0.9\linewidth]{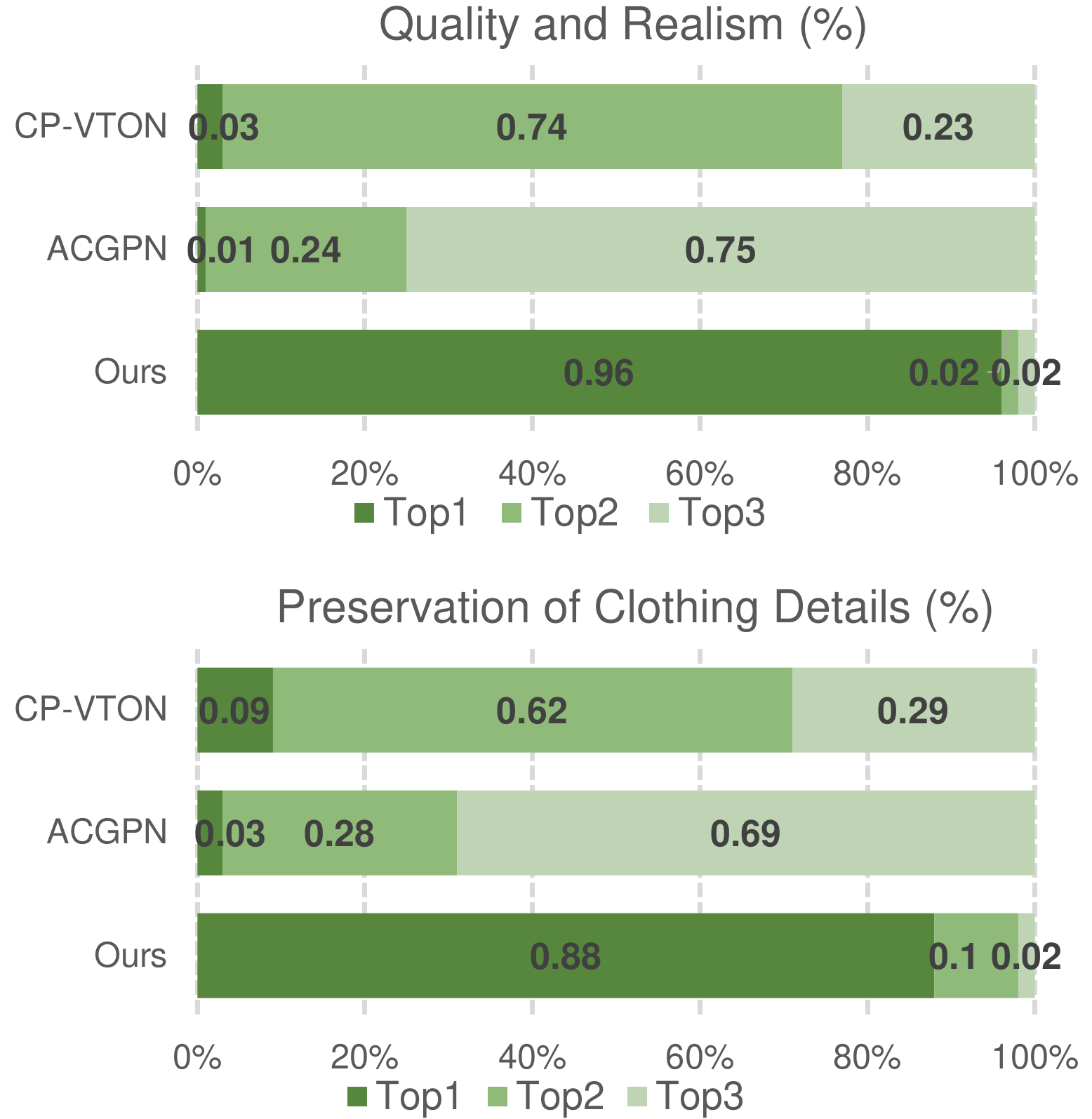}
    \vspace{-0.1cm}
    \caption{User study results. We compare our model with CP-VTON~\cite{wang2018toward} and ACGPN~\cite{yang2020towards}.}
    \label{fig:experiment-userstudy}
\end{figure}

\begin{figure*}[t!]
    \centering
    \includegraphics[width=0.95\linewidth]{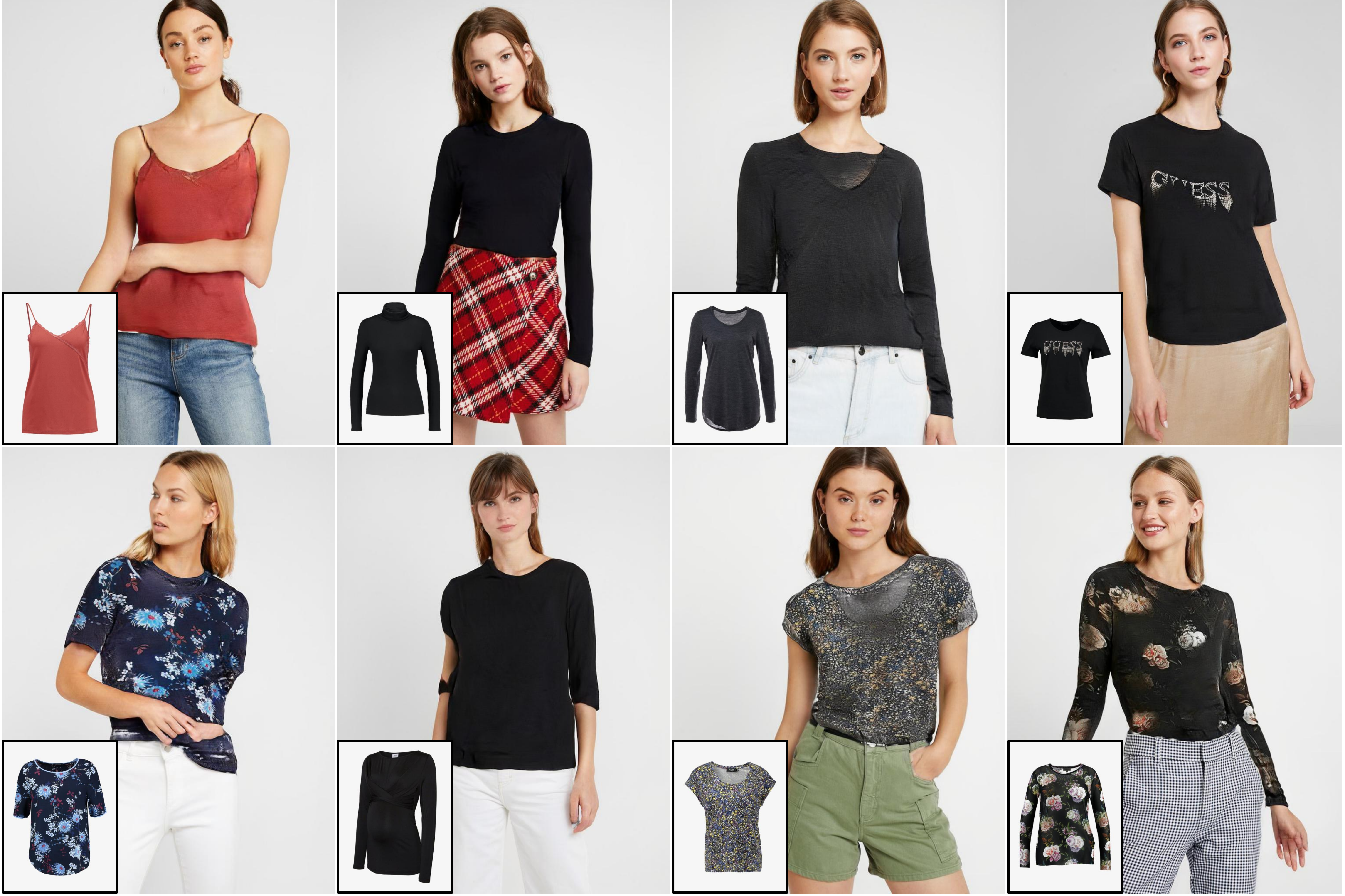}
    \caption{Failure cases of VITON-HD.}
    \label{failure_cases}
\end{figure*}

\subsection*{B.2. User Study}
We further evaluate our model and other baselines via a user study in the unpaired setting.
We randomly select 30 sets of a reference image and a target clothing image from the test dataset.
Given the reference images and the target clothes, the users are asked to rank the 1024$\times$768 outputs of our model and baselines according to the following questions:
(1) Which image is the most photo-realistic?
(2) Which image preserves the details of the target clothing the most?
As shown in Fig.~\ref{fig:experiment-userstudy}, it can be observed that our approach achieves the rank 1 votes more than 88\% for the both questions.
The result demonstrates that our model generates more realistic images, and preserves the details of the clothing items compared to the baselines.

\subsection*{B.3. Qualitative Results}
\begin{figure*}[t!]
    \centering
    \includegraphics[width=\linewidth]{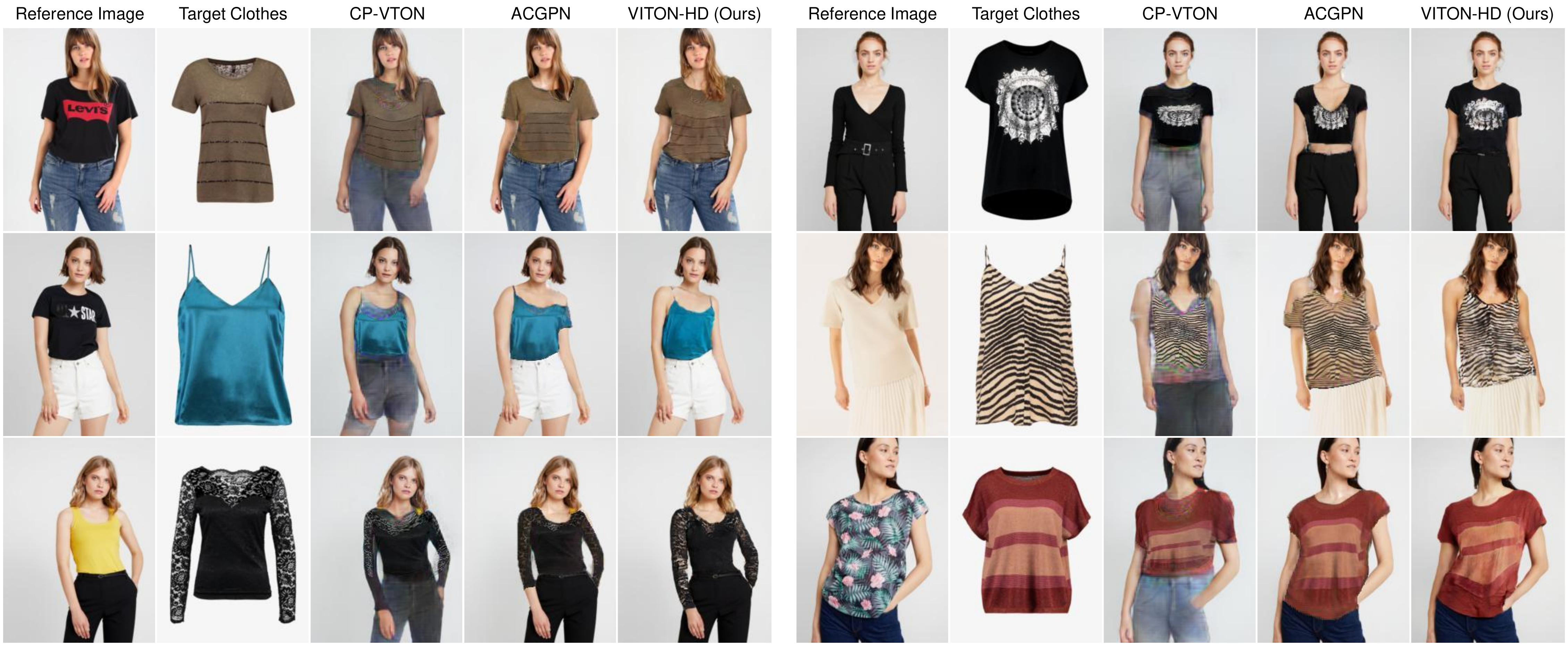}
    \caption{Qualitative comparison of the baselines (256$\times$192).}
    \label{fig:supp-256x192}
\end{figure*}

\begin{figure*}[t!]
    \centering
    \vspace{-0.015cm}
    \includegraphics[width=0.7\linewidth]{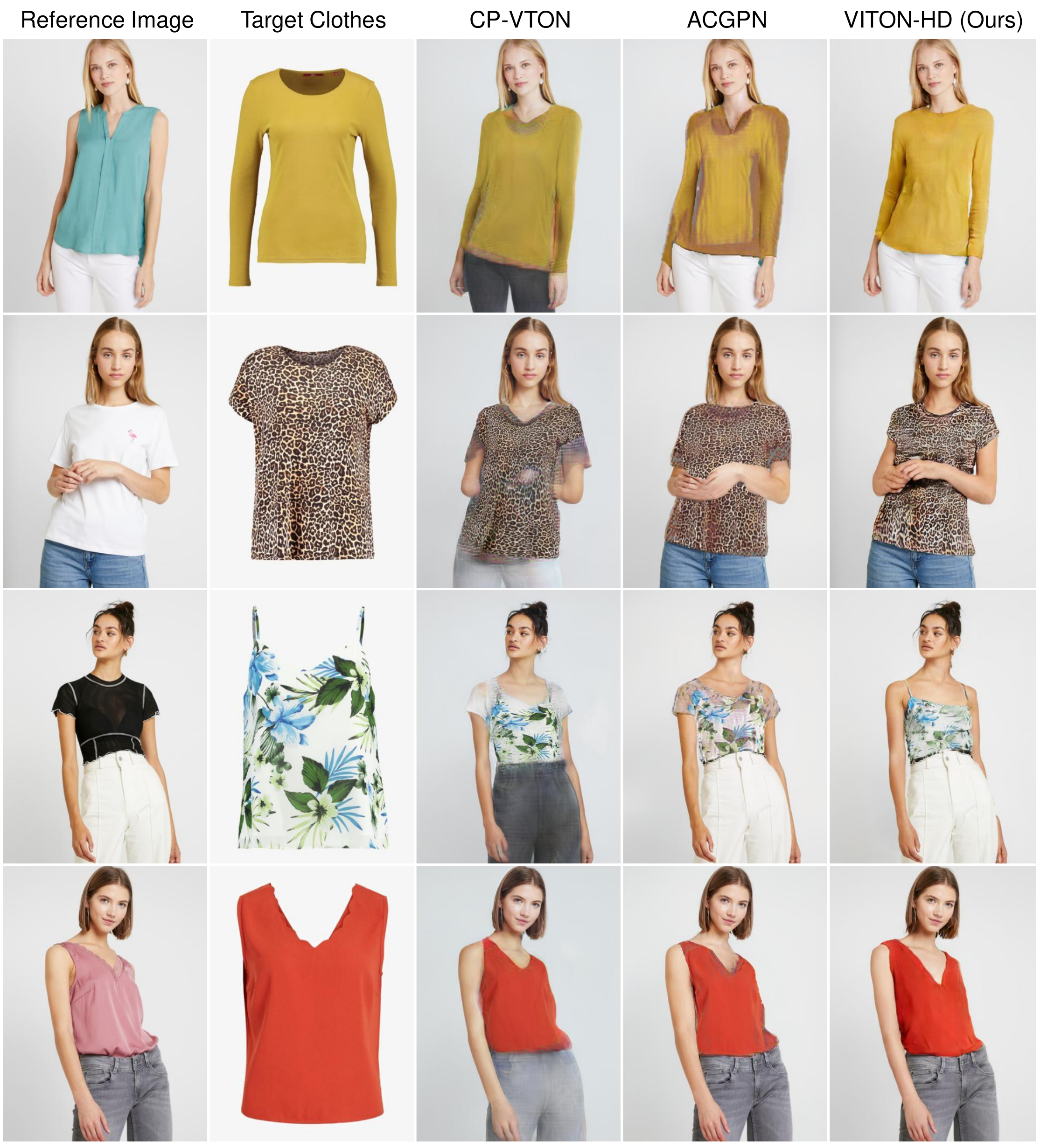}
    \caption{Qualitative comparison of the baselines (512$\times$384).}
    \label{fig:supp-512x384}
\end{figure*}

\begin{figure*}[t!]
    \centering
    \includegraphics[width=\linewidth]{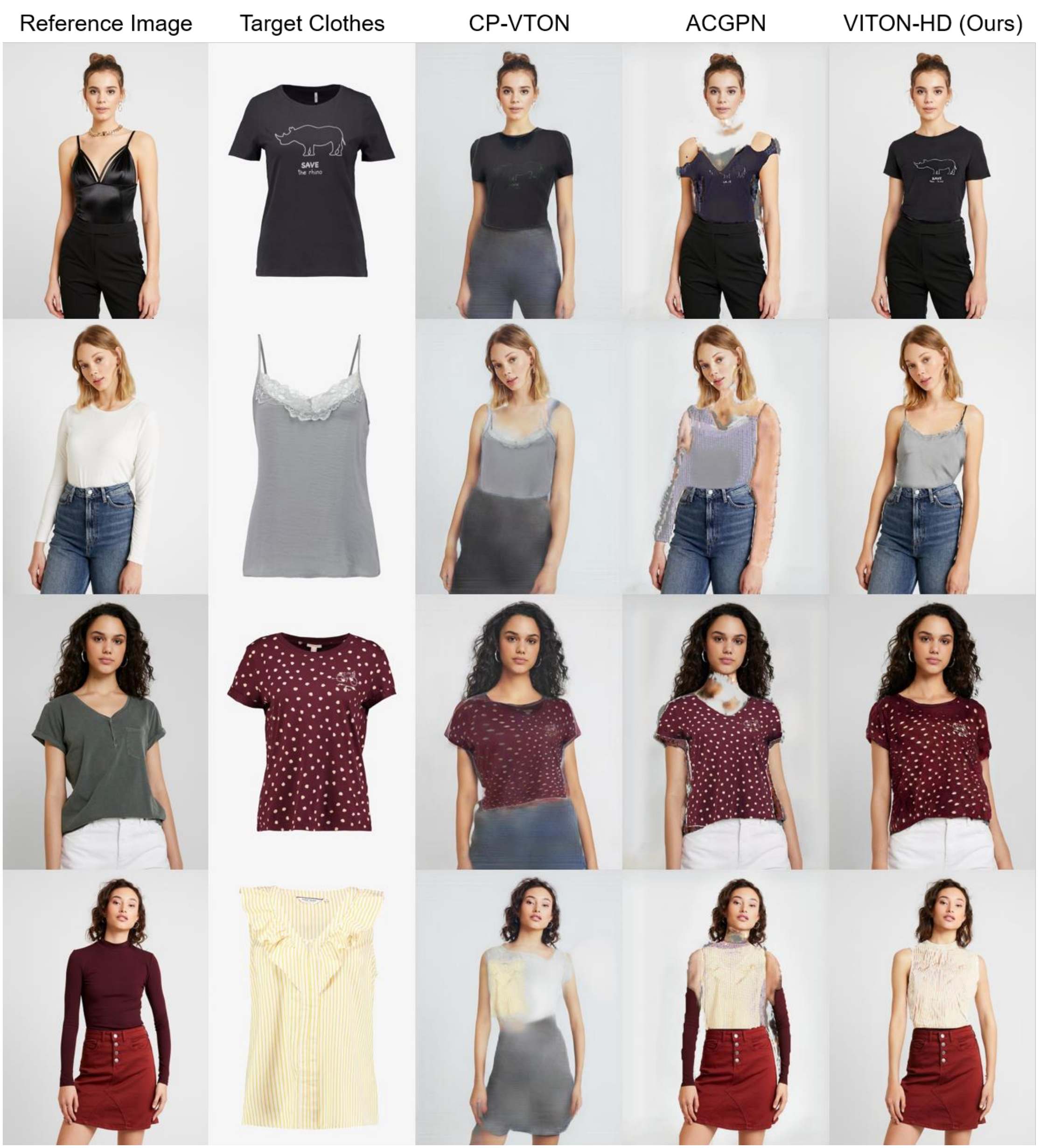}
    \caption{Qualitative comparison of the baselines (1024$\times$768).}
    \label{fig:supp-1024x768_1}
\end{figure*}

\begin{figure*}[t!]
    \centering
    \includegraphics[width=\linewidth]{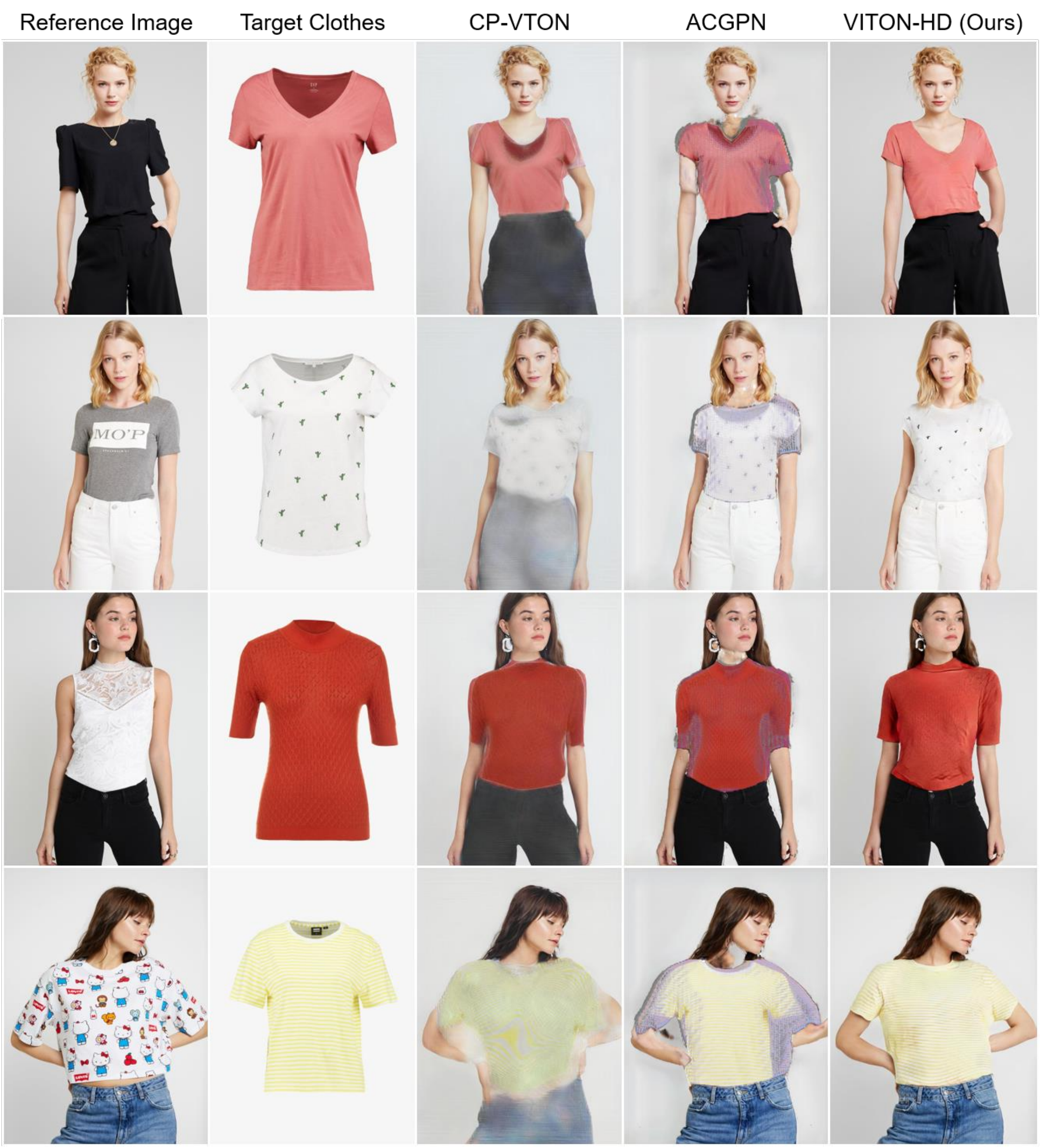}
    \caption{Qualitative comparison of the baselines (1024$\times$768).}
    \label{fig:supp-1024x768_2}
\end{figure*}

\begin{figure*}[t!]
    \centering
    \includegraphics[width=0.95\linewidth]{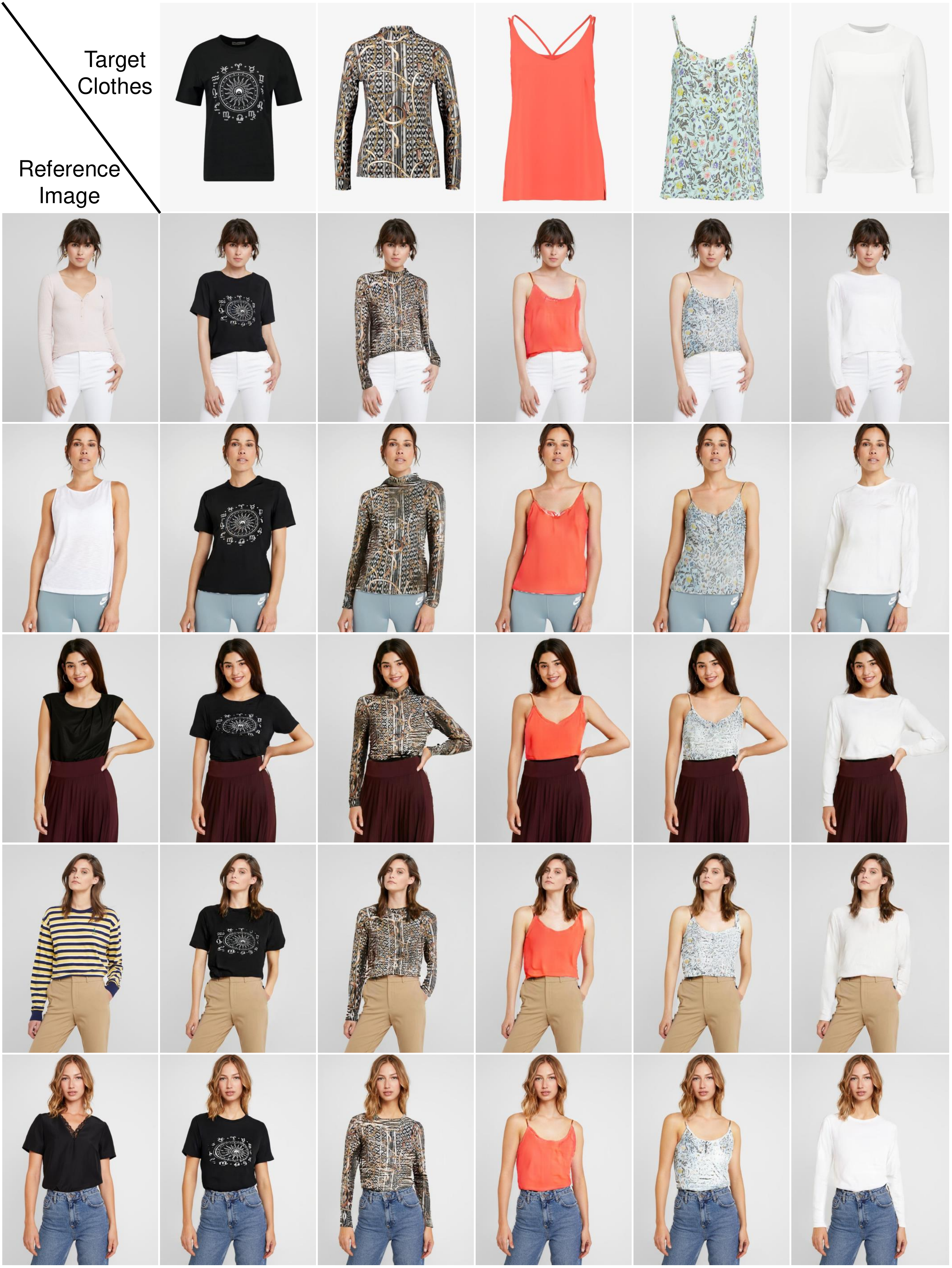}
    \caption{Additional qualitative results of \model.}
    \label{fig:supp-matrix}
\end{figure*}

\begin{figure*}[t!]
    \centering
    \includegraphics[width=\linewidth]{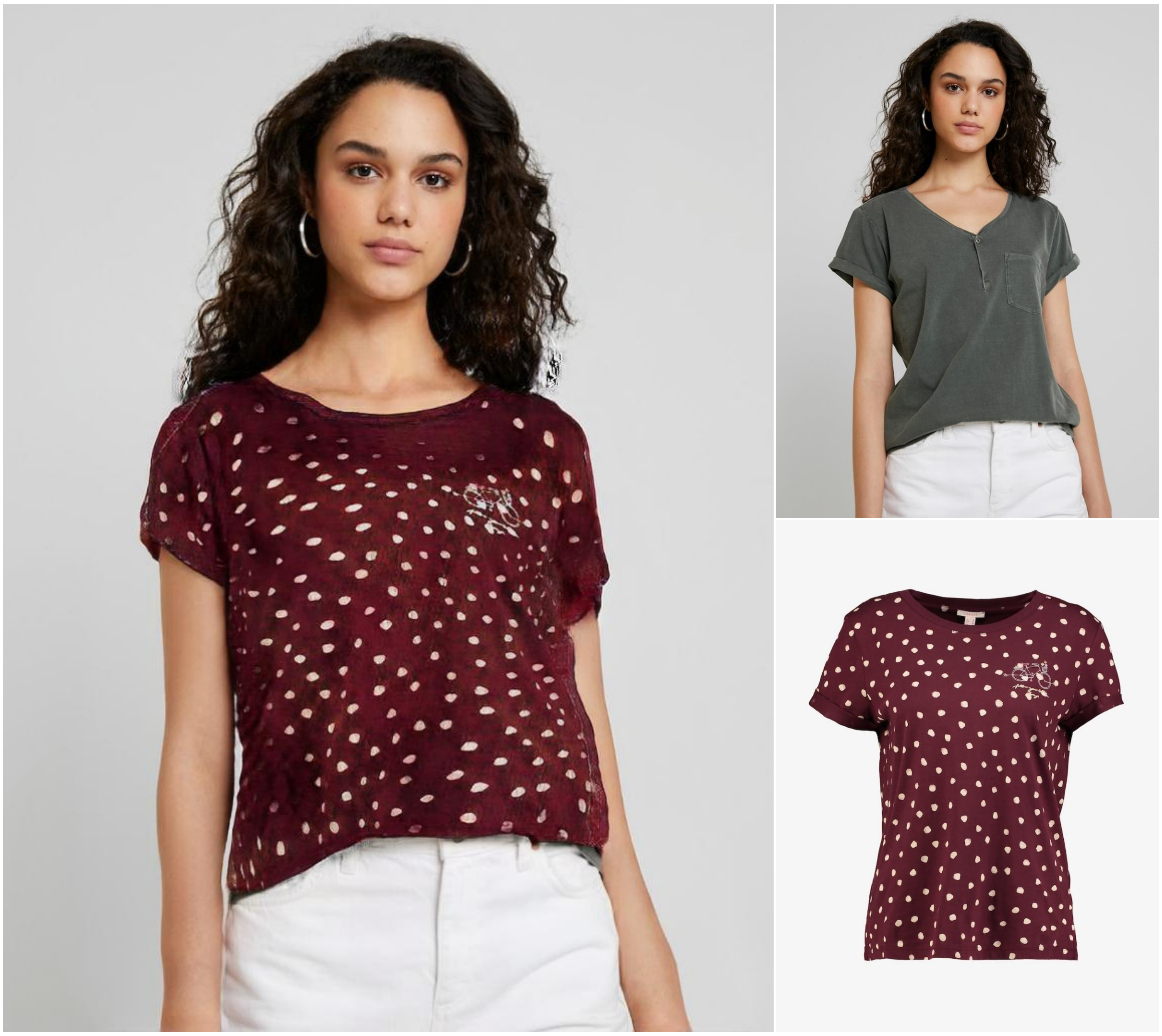}
    \caption{Sample 1 of \model. (\textit{Left}) The synthetic image. (\textit{Right}) The reference image and the target clothing item.}
    \label{fig:supp-case1}
\end{figure*}

\begin{figure*}[t!]
    \centering
    \includegraphics[width=\linewidth]{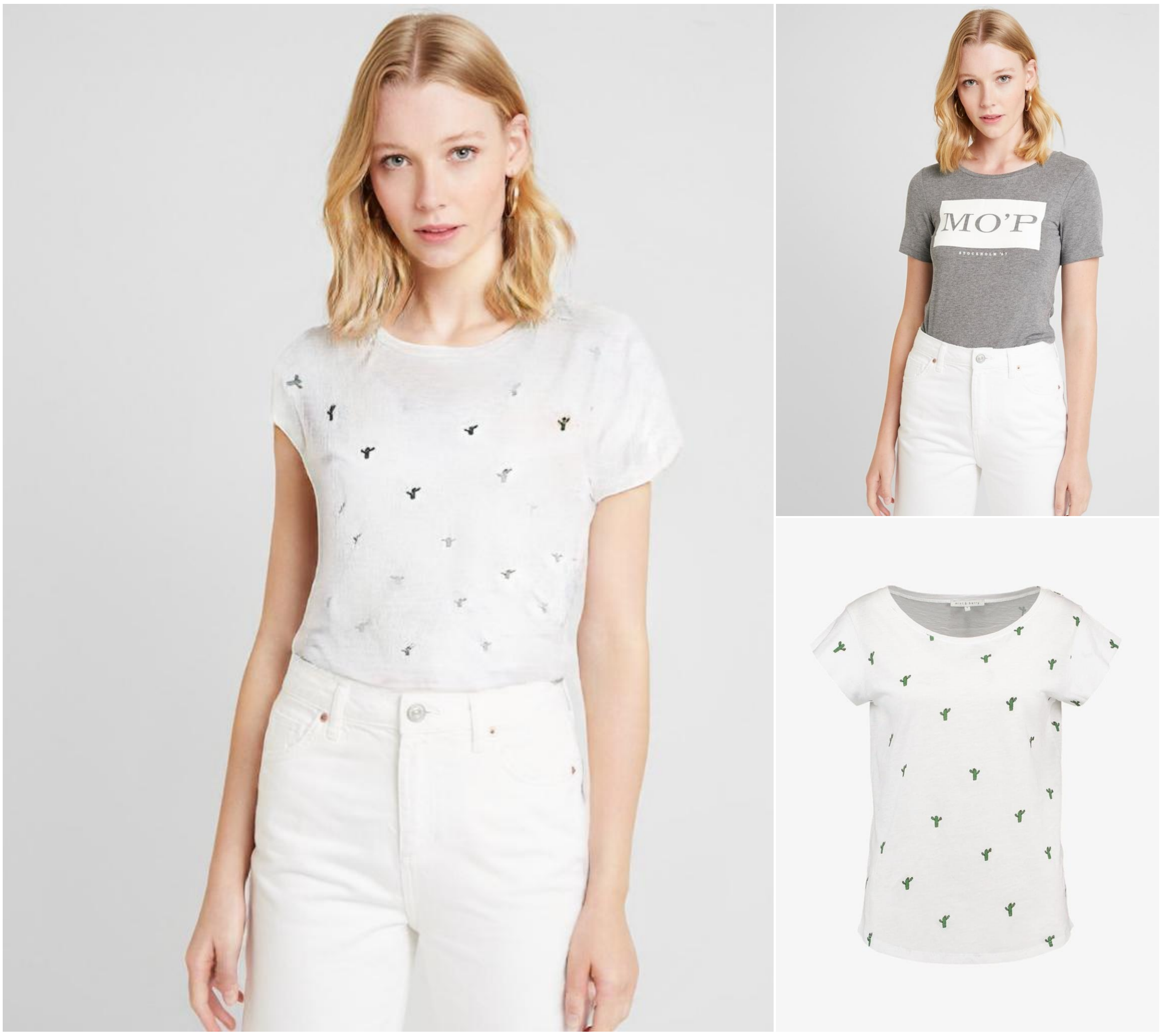}
    \caption{Sample 2 of \model. (\textit{Left}) The synthetic image. (\textit{Right}) The reference image and the target clothing item.}
    \label{fig:supp-case2}
\end{figure*}

\begin{figure*}[t!]
    \centering
    \includegraphics[width=\linewidth]{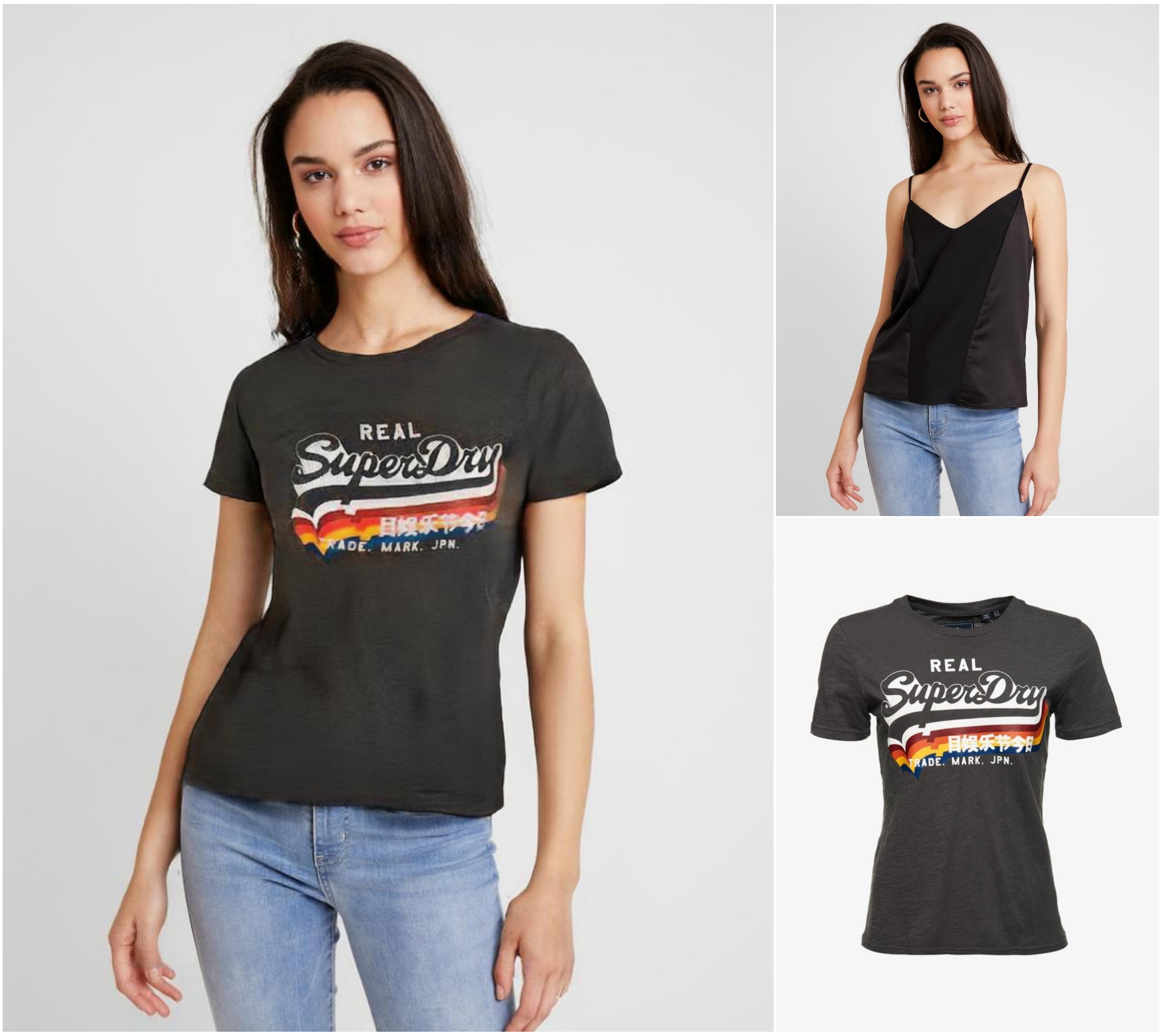}
    \caption{Sample 3 of \model. (\textit{Left}) The synthetic image. (\textit{Right}) The reference image and the target clothing item.}
    \label{fig:supp-case3}
\end{figure*}

We provide additional qualitative results to demonstrate our model's capability of handling high quality image synthesis. Fig.~\ref{fig:supp-256x192}, ~\ref{fig:supp-512x384}, ~\ref{fig:supp-1024x768_1}, and ~\ref{fig:supp-1024x768_2} show the qualitative comparison of the baselines across different resolutions.
Fig.~\ref{fig:supp-matrix}, ~\ref{fig:supp-case1}, ~\ref{fig:supp-case2}, and ~\ref{fig:supp-case3} show additional results of VITON-HD at 1024$\times$768 resolution.

\section*{C. Failure Cases and Limitations}
Fig.~\ref{failure_cases} shows the failure cases of our model caused by the inaccurately predicted segmentation map or the inner collar region indistinguishable from the other clothing region.
Also, the boundaries of the clothing textures occasionally fade away.

The limitations of our model are as follows.
VITON-HD is trained to preserve the bottom clothing items, limiting the presentation of the target clothes (\eg, whether they are tucked in).
It can be a valuable future direction to generate multiple possible outputs from a single input pair.
Next, our dataset mostly consists of slim women and top clothing images, which makes VITON-HD handle only a limited range of body shapes and clothing during the inference.
We believe that VITON-HD has the capability to cover more diverse cases when the images of various body shapes and clothing types are provided.
Finally, existing virtual try-on methods including VITON-HD do not provide robust performance for in-the-wild images.
We think generating realistic try-on images for the in-the-wild images is an interesting topic for future work.

\clearpage

\end{document}